\documentclass[12pt]{elsarticle}

\usepackage{amssymb}
\usepackage{amsmath}
\usepackage{subfig}
\usepackage{graphicx}
\usepackage{caption}
\usepackage{tikz}
\usepackage{makecell}
\usepackage{hyperref}
\usepackage{multirow}
\usepackage{rotating}

\journal{ }

\begin{document}

\begin{frontmatter}



\title{OmniAD: View-Invariant Unsupervised Scene Pixelwise Anomaly Detection with Adaptive View Synthesis}

\author[1]{Subin Varghese}
\author[1,2]{Vedhus Hoskere}

\affiliation[1]{
    organization={Department of Electrical and Computer Engineering, University of Houston}, 
    addressline={4226 Martin Luther King Blvd.}, 
    city={Houston},
    state={Texas},
    postcode={77204},
    country={USA}
}

\affiliation[2]{
    organization={Department of Civil and Environmental Engineering, University of Houston}, 
    addressline={4226 Martin Luther King Blvd.}, 
    city={Houston},
    state={Texas},
    postcode={77204},
    country={USA}
}

\begin{abstract}
The built environment, encompassing critical infrastructure such as bridges and buildings, requires diligent monitoring of unexpected anomalies or deviations from a normal state in captured imagery. Anomaly detection methods could aid in automating this task; however, deploying anomaly detection effectively in such environments presents significant challenges that have not been evaluated before. These challenges include camera viewpoints that vary, the presence of multiple objects within a scene, and the absence of labeled anomaly data for training. To address these comprehensively, we introduce and formalize Scene Anomaly Detection (Scene AD) as the task of unsupervised, pixel-wise anomaly localization under these specific real-world conditions. Evaluating progress in Scene AD required the development of ToyCity, the first multi-object, multi-view real-image dataset, for unsupervised anomaly detection. Our initial evaluations using ToyCity revealed that established anomaly detection baselines struggle to achieve robust pixel-level localization. To address this, two data augmentation strategies were created to generate additional synthetic images of non-anomalous regions to enhance generalizability. However, the addition of these synthetic images alone only provided minor improvements. Thus, OmniAD, a refinement of the Reverse Distillation methodology, was created to establish a stronger baseline. Our experiments demonstrate that OmniAD, when used with augmented views, yields a 64.33\% increase in pixel-wise \(F_1\) score over Reverse Distillation with no augmentation. Collectively, this work offers the Scene AD task definition, the ToyCity benchmark, the view synthesis augmentation approaches, and the OmniAD method. Project Page: https://drags99.github.io/OmniAD/
\end{abstract}




\end{frontmatter}



\section{Introduction}
\label{sec:intro}

Detecting anomalies is critical for many scene assessment and analysis applications in the built environment, including infrastructure inspections\cite{Hoskere2019Vision-BasedVehicles,Hoskere2025UnifiedFF,Rakoczy2024TechnologiesAP}, disaster response\cite{Singh2024MulticlassPB,Singh2023PostDD}, construction management \cite{Gu2024DigitalTA} and planning
\cite{Li2025ARO}.
Anomalies can be described as a deviation (e.g., damage, structural changes, new objects) from an established non-anomalous reference. In many scene assessment and analysis applications, images are the primary sensing modality, as they are inexpensive to capture and provide the dense resolution required for visual evidence in documentation \cite{txdotINS,pu2024automated}. 


Anomaly detection (AD) approaches could be categorized as supervised or unsupervised. Supervised approaches rely on annotated data and work well when high-quality annotations are available but can be impractical in many real-world inspection scenarios where anomalies are rare or unknown in advance \cite{Bergmann2019MVTECDetection,Zou2022SPot-the-DifferenceAndSegmentation,ZhouPad:Detection}. Unsupervised AD methods, on the other hand, are often more practical as they don't require labeled data of anomalies. Current benchmarks for unsupervised anomaly detection have largely been tailored for manufacturing contexts which consist of imagery of a single-object \cite{bergmann2019mvtec,silvestre2019public,ZhouPad:Detection} or multiple objects \cite{zou2022spot,bergmann2022beyond} with constrained views.

\begin{figure}
    \centering
    \subfloat[]{\includegraphics[width=3in]{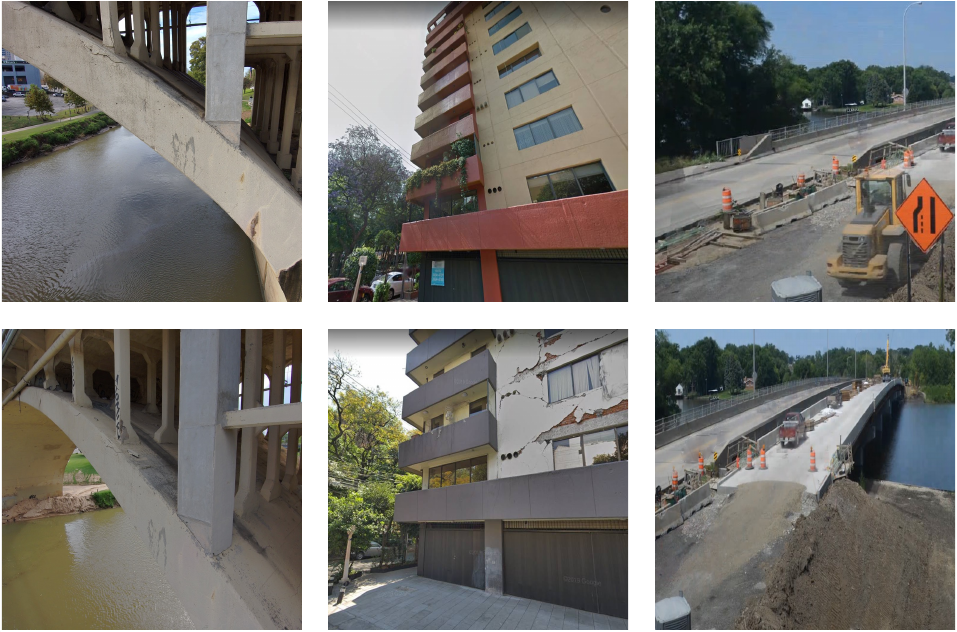}%
    \label{fig_inspection}}
    \hfil
    \subfloat[]{\includegraphics[width=3in]{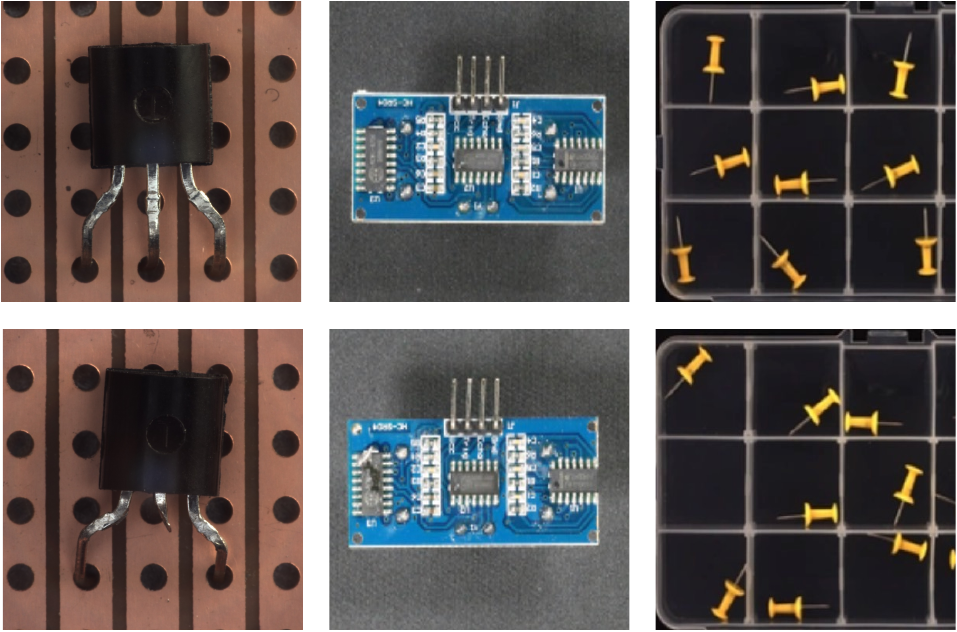}
    \label{fig_mvtec}}
    \caption{(a) Typical images used in the built environment. (b) Images available for benchmarking unsupervised anomaly detection \cite{bergmann2019mvtec,zou2022spot,bergmann2022beyond}. The top row for (a) and (b) consist of non-anomalous images, while the bottom is anomalous.}
    \label{fig:inspection_v_mvtec}
\end{figure}

While unsupervised image-based AD methods have found great success in manufacturing applications, they have not been explored for scene analysis applications in the built environment, where many manufacturing-specific assumptions no longer hold. Imagery in the built environment is typically collected from unconstrained viewpoints, often by handheld or UAV-mounted cameras \cite{Hoskere2019Vision-BasedVehicles,Varghese2024ViewDeltaTC,Narazaki2022Vision-basedVehicles,Varghese2023UnpairedIT}. As shown in \figurename~\ref{fig:inspection_v_mvtec} view-points for images captured in the built environment vary in nature tremendously, unlike the fixed-camera setting typically used for unsupervised AD benchmarks \cite{bergmann2019mvtec,ZhouPad:Detection,jezek2021deep,zou2022spot}. Seminal anomaly detection approaches \cite{Gudovskiy2022CFLOW-AD:Flows,Yu2021FastFlow:Flows,Lee2022Cfa:Localization,Zavrtanik2021DRMDetection,Deng2022AnomalyEmbedding} discussed in more detail in the related works section have matured to the point of achieving very high quantitative metrics with fixed views, with state-of-the-art methods plateauing in performance. However, our studies showed that seminal works are unable to meet similar performance levels for multi-view multi-object scenes such as in \figurename~\ref{fig:inspection_v_mvtec} (a). This performance disparity underscores the need to specifically define and address the problem of anomaly detection within these unconstrained visual environments.

\begin{figure}
    \centering
    \includegraphics[width=.9\linewidth]{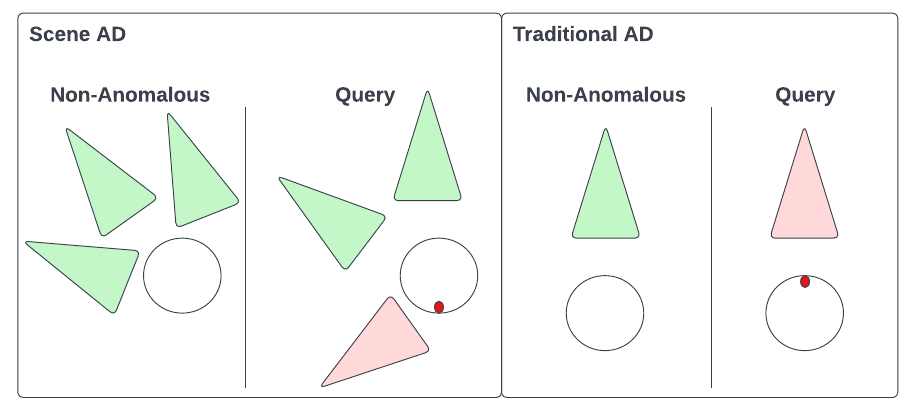}
    \caption{In Scene AD, we observe variations in camera positions between non-anomalous views (green) and query views that may contain anomalies (red). In contrast, traditional AD methods typically feature aligned views between the non-anomalous and anomalous images.}
    \label{fig:Scene_AD}
\end{figure}

An effective scene anomaly detection method in unconstrained visual environments must operate (i) without labeled data, (ii) across multiple objects per image, and (iii) under random viewpoint variation. We formally refer to this general AD task as \textbf{Scene Anomaly Detection (Scene AD)} and illustrate it in \figurename~\ref{fig:Scene_AD}. Formally, Scene AD entails localizing anomalous regions given two unlabeled image sets: one depicting normal reference conditions, called the non-anomalous set, and one containing potential anomalies, if any, called the query set. No assumption is made for alignment between the non-anomalous and query set. Camera poses, if needed, should only be estimates from methods such as structure-from-motion or GPS measurements to mimic the noisy camera poses that can feasibly be obtained in an uncontrolled environment.

      


To provide a benchmark for Scene AD, we introduce \textbf{ToyCity} a real-world dataset consisting of four multi-object scenes, with more than 1200 anomalous images per scene and over 20 varying anomalies across scenes. The diversity of anomalies and quantity of anomalous images in ToyCity match or exceed those of the most frequently used fixed view AD benchmarks \cite{zou2022spot, carrera2016defect, bergmann2022beyond, bergmann2021mvtec, bergmann2019mvtec, jezek2021deep, silvestre2019public}. ToyCity also exceeds the real-image diversity of the Multi-Pose Anomaly Detection (MAD)\cite{ZhouPad:Detection} dataset, which contains 16 to 32 anomalous images per object. When evaluating baseline AD methods, that perform well on prior datasets, we found they exhibit very weak pixel-level localization on ToyCity. Among the baseline AD methods, Reverse Distillation (RD) \cite{Deng2022AnomalyEmbedding} showed the best localization performance, however we found its localization performance to be inconsistent for practical use.


We introduce \textbf{OmniAD} to validate the potential of unsupervised AD approaches for  pixel-level Scene Anomaly Detection. OmniAD is a refinement of Reverse Distillation \cite{Deng2022AnomalyEmbedding} that integrates \textbf{student-attention modules}. These modules are designed to expand the student decoder's effective receptive field (ERF) \cite{Liu2018UnderstandingSegmentation}, thereby improving its sensitivity to localized anomalies. Furthermore, to counteract the performance degradation caused by viewpoint variations in Scene AD, we explore two data augmentation strategies. Both strategies aim to generate novel non-anomalous views, using Neural Radiance Fields (NeRF) \cite{Mildenhall2020NeRF:Synthesis} and hierarchical localization (hloc) \cite{Detone2018SuperPoint:Description,Sarlin2020SuperGlue:Networks}, to assist AD methods in generalizing to varying views.

Our main contributions are:
\begin{itemize}
    \item \textbf{Scene AD Task}: Formal introduction of Scene Anomaly Detection, highlighting the challenge of detecting anomalies in unaligned multi-view imagery consisting of multiple objects.
    \item \textbf{ToyCity Dataset}: A new multi-object multi-view dataset of real images designed for benchmarking methods for Scene AD.
    \item \textbf{Baselines:} We establish the first quantitative metrics of seminal AD methods for Scene AD using ToyCity.
    \item \textbf{NeRF based Data Augmentation Methods}: Two augmentation strategies for increased view-generalization in Scene AD.
    \item \textbf{OmniAD}: A refinement of Reverse Distillation \cite{Deng2022AnomalyEmbedding}, which when combined with augmentation strategies leads to a 64.33\% improvement in pixel-wise \(F_1\) score for Scene AD compared to Reverse Distillation with no augmentation.

\end{itemize}

The rest of the paper is organized as follows:  
Section~\ref{sec:related_work} reviews related work on anomaly detection, neural radiance fields, and camera localization.  
Section~\ref{sec:method} introduces the proposed OmniAD architecture and novel data augmentation strategies.  
Section~\ref{sec:dataset} presents the ToyCity dataset.  
Section~\ref{sec:experiments} details the experimental setup, metrics, and results.  
Section~\ref{sec:limitations} discusses the limitations of our dataset and method.  
Finally, Section~\ref{sec:conclusion} summarizes and concludes this paper.

\section{Related Work}
\label{sec:related_work}
\subsection{Unsupervised Image-Based Anomaly Detection}
Anomaly detection (AD) algorithms typically harness a collection of non-anomalous images, to establish a normative distribution. AD is a critical function in industrial visual inspection for identifying defects \cite{Bergmann2019MVTECDetection} and is also employed in healthcare for the detection of irregularities in medical images \cite{Schlegl2019F-AnoGAN:Networks}. Unsupervised AD methods have demonstrated considerable success, particularly for fixed-view single-object anomaly detection. This domain has become mature over time to achieve a high-performance plateau for pixel-wise anomaly detection. New methods \cite{Kang2024MSTADAM,He2024MambaADES,Wang2024WeaklySA,Yao2024PriorNP} frequently benchmark against a core group of foundational techniques \cite{Gudovskiy2022CFLOW-AD:Flows,Yu2021FastFlow:Flows,Lee2022Cfa:Localization,Zavrtanik2021DRMDetection,Deng2022AnomalyEmbedding} due to the high-performance plateau achieved.

When introducing novel tasks such as Scene AD, we must select robust baseline methods that are demonstrably reproducible and widely accepted for comparison to create a meaningful benchmark. Accordingly, our analysis centers on prominent methods such as CFLOW-AD \cite{Gudovskiy2022CFLOW-AD:Flows}, FastFlow \cite{Yu2021FastFlow:Flows}, Coupled-hypersphere-based Feature Adaptation (CFA) \cite{Lee2022Cfa:Localization}, discriminatively trained reconstruction anomaly embedding model (DRÆM) \cite{Zavrtanik2021DRMDetection}, and Reverse Distillation (RD) \cite{Deng2022AnomalyEmbedding}. These specific methods cover the two primary categories unsupervised image-based AD methods fall into, feature embedding based and reconstruction based \cite{ZhouPad:Detection, He2024MambaADES}. Furthermore, this strategic selection aims to establish a durable and widely understood performance benchmark. These methods offer strong guarantees of reproducibility, stemming from their continuous validation and the availability of robust, often multiple, implementations \cite{anomalib}. 

\subsection{Feature Embedding Based Unsupervised Anomaly Detection}
Both CFLOW-AD and FastFlow utilize a flow framework \cite{Dinh2017DensityNVP} that models the distribution of normal image patches to detect anomalous patterns. Typically, these methods begin by extracting features from a pretrained CNN encoder for an input image patch. The feature vectors are then mapped to a multivariate Gaussian distribution through the flow framework. Anomalous images can then be detected because they tend to fit poorly to this learned distribution, highlighting their abnormality.

The CFA method uses a combination of a frozen pretrained CNN, a patch descriptor network that learns on-the-fly, and a memory bank for storing compressed features. CFA fine-tunes the pretrained CNN features through metric learning, which is based on a coupled hypersphere approach, clustering normal features tightly within a hypersphere's boundary.

\subsection{Reconstruction Based Unsupervised Anomaly Detection}

DRÆM operates with two main networks: a reconstructive network and a discriminative network. DRÆM generates synthetic anomalies that the reconstructive network then transforms back to normal. The discriminative network uses both the reconstructed normal image and the synthetic anomalous image to predict segmentation for the regions where the synthetic anomaly was introduced.

Reverse Distillation adopts a knowledge distillation strategy \cite{Bergmann2020UninformedEmbeddings,Salehi2021MultiresolutionDetection}, where a student decoder network learns the feature space representation from a teacher encoder network at various feature map scales. Unlike traditional approaches that feed the input image into both the student and the teacher networks, Reverse Distillation takes an autoencoder-like approach where the teacher encodes the input image, and the student decodes a fused multiscale feature map output by the teacher. Deviations between the student and teachers' features indicate the occurrence of an anomaly.

\subsection{Multi-View Anomaly Detection}
While the aforementioned unsupervised AD methods demonstrate strong performance in controlled environments, a significant limitation inherent in many of these 2D approaches arises when deployed in multi-view settings \cite{ZhouPad:Detection}. Their effectiveness often depends critically on feature representations or learned normality models (such as distributions, reconstruction targets, or feature manifolds) that are derived from training data \cite{Bergmann2019MVTECDetection, Zou2022SPot-the-DifferenceAndSegmentation}. As a result, these methods frequently struggle when presented with the significant viewpoint variations from those seen in the training data that is typical of real-world data acquisition, like images captured using handheld devices or UAVs during infrastructure assessments \cite{Hoskere2019Vision-BasedVehicles, Varghese2023UnpairedIT}. Additionally, realistic scenes often contain multiple objects simultaneously, a scenario largely distinct from typical single-object AD benchmarks. Consequently, the specific task of unsupervised anomaly detection within environments characterized by both multiple objects per scene and random, unconstrained multi-view captures has remained largely unexplored. This is partly due to the lack of suitable benchmark datasets designed to address these combined complexities, as the manual effort in creating such a dataset with real images is tremendous in comparison to labeling the occurrence of an anomaly on a single object.


\subsection{Neural Radiance Field}
Neural Radiance Fields (NeRFs) \cite{Mildenhall2020NeRF:Synthesis} represent a significant advancement in the field of view synthesis. These fields use a collection of images with known camera poses to synthesize new viewpoints of a scene. Unlike conventional 3D Structure-from-Motion (SfM) methods that construct explicit geometric models, NeRFs model a scene's volume through a neural network that predicts color and density for space points along camera rays. This technique enables the creation of photorealistic images from novel perspectives. Subsequent innovations like Nerfacto \cite{Tancik2023Nerfstudio:Development} and Instant-ngp \cite{Muller2022InstantEncoding} have further refined NeRFs, offering rapid training and real-time rendering capabilities. In the context of this work, we leverage NeRFs to synthesize non-anomalous images from diverse viewpoints.

\subsection{Hierarchical Localization}
The process of hierarchical localization (hloc) \cite{Sarlin2019FromScale, Sarlin2020SuperGlue:Networks} has been developed to accurately determine a camera's position and orientation in space, a task known as 6-Degree-of-Freedom localization \cite{Sarlin2019FromScale,Sarlin2020SuperGlue:Networks,LvUnbiasedDetection}. This process begins with the extraction of local features from a query image, utilizing methods such as SuperPoint \cite{Detone2018SuperPoint:Description}, Sift \cite{Lowe2004DistinctiveKeypoints}, or DISK \cite{Tyszkiewicz2020DISK:Gradient}. A 3D SfM model serves as a reference for these local features, against which the query image is matched using global descriptors like VLAD \cite{Goh2014LearningCoding}. The Perspective-n-Point algorithm, with RANSAC, then utilizes these correspondences to estimate the camera's pose. This pose is critical in enabling NeRFs to generate non-anomalous images that correspond to those containing anomalies, thus in theory providing a more robust dataset for anomaly detection algorithms. However, as we shall show, an improved AD method is needed to fully utilize these anomaly aligned images.

\section{Method}
\label{sec:method}

\begin{figure}
    \centering
    \includegraphics[width=\linewidth]{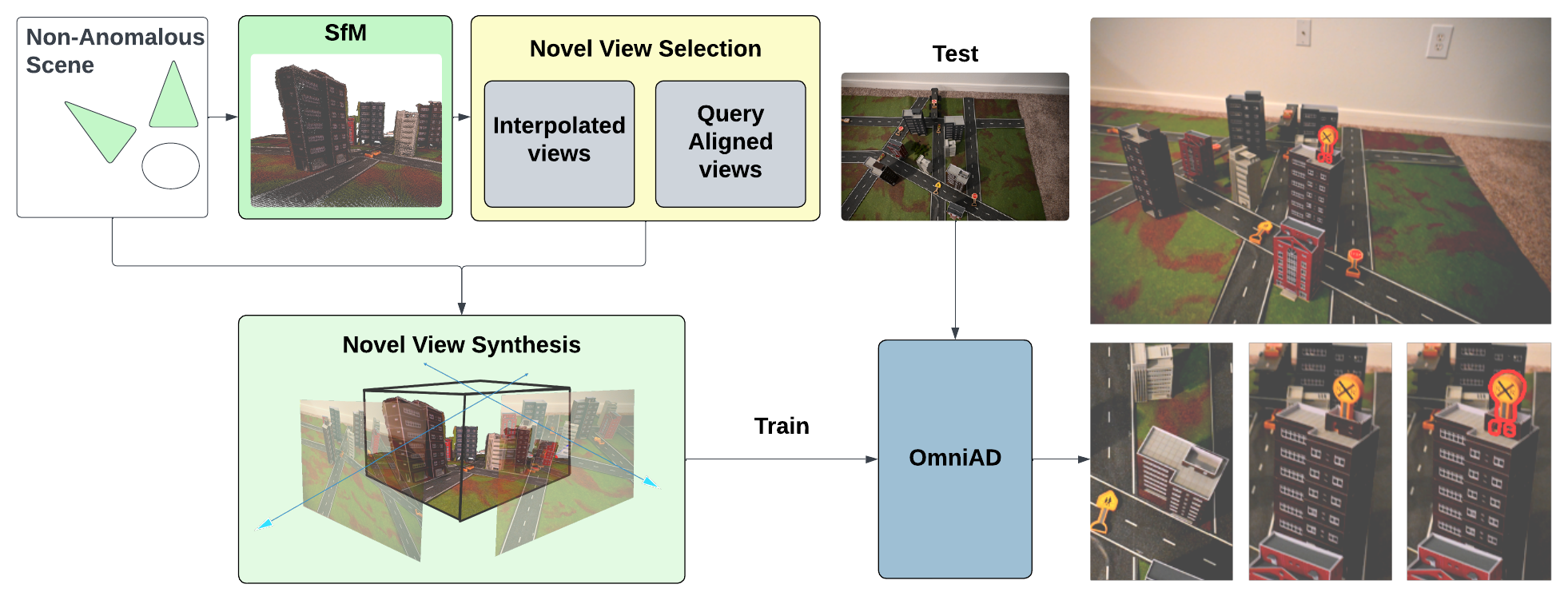}

    \caption{Overview of our methodology is depicted here. We utilize the non-anomalous views from a Scene AD to generate a non-anomalous SfM model. A novel view selection strategy utilizes anomaly views and the SfM model to generate novel views to augment the non-anomalous dataset. OmniAD is then trained on the augmented non-anomalous dataset for anomaly detection.}
    \label{fig:overview}
\end{figure}

\begin{figure}
    \centering
    \includegraphics[width=\linewidth]{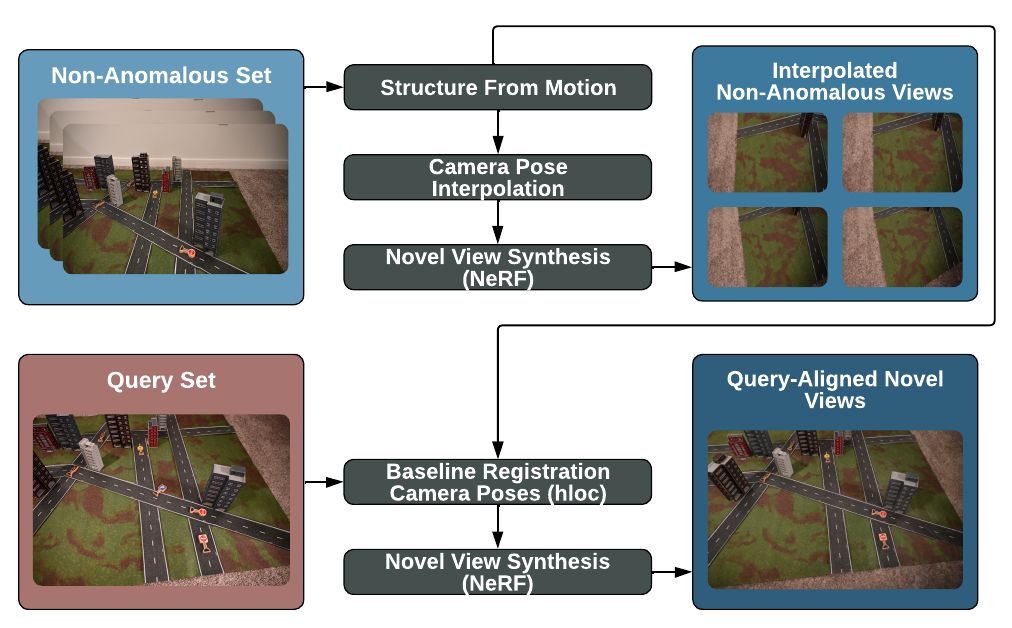}
    \caption{Overview of our novel view selection strategy. Utilizing the non-anomalous dataset, we generate a structure from motion (SfM) 3D model and accompanying camera pose for each image. Leveraging these non-anomalous camera poses, we design a trajectory that connects them, allowing for interpolation. Subsequently, we utilize a NeRF for Novel View Synthesis (NVS) to generate images that present unique perspectives along the interpolated path. Query images can also be used to determine more novel views.}
    \label{fig:nerf_augmentation_overview}
\end{figure}

An overview of our AD methodology for Scene AD can be seen in \figurename~\ref{fig:overview}. Scene AD consists of non-anomalous views and query views. We generate an SfM model using non-anomalous views to acquire camera poses. Leveraging these non-anomalous camera poses, we can implement two different novel view selection strategies to produce images for data augmentation, as shown in \figurename~\ref{fig:nerf_augmentation_overview}. For the first strategy, we interpolate between the non-anomalous camera poses. The second strategy for generating useful camera views for training is to utilize hloc to acquire the camera poses of query views. We then use a NeRF for novel view synthesis (NVS) with the new camera poses to augment our non-anomalous views for training our proposed method OmniAD. The following subsections describe each of these steps in more detail. 

\subsection{Novel View Selection Strategy}
A core challenge in Scene AD is the viewpoint variation between the non-anomalous reference images and the query images. Anomaly detection models trained only on the original non-anomalous views may struggle when presented with a query image from a significantly different perspective. To enhance the model's robustness to these variations, a wider variety of non-anomalous viewpoints are required for training, ideally including views similar to those encountered in the query set.

To generate these new non-anomalous camera poses for novel view synthesis (NVS), we propose and evaluate two different approaches:
\begin{itemize}
    \item \textbf{Interpolated Non-anomalous Views (INV):} The goal of this strategy is to densify the coverage of the non-anomalous viewpoint space. By generating views between the existing captured non-anomalous poses, we train the model to handle gradual shifts in perspective, improving its general robustness to unseen viewpoints near the original capture trajectory.
    \item \textbf{Query-Aligned Non-anomalous Views (QANV):} This strategy directly aims to address the misalignment with potentially anomalous query images. By estimating the camera pose of a query image and then synthesizing a non-anomalous view from that approximate camera pose, we provide the model with similar normal/potentially anomalous pairs (in terms of viewpoint). Intuitively, this allows the model to better isolate deviations caused by anomalies rather than viewpoint changes.
\end{itemize}
The following subsections detail how these camera poses are generated and used with NVS.

\subsubsection{Interpolated Non-anomalous Views (INV)}

\begin{figure}
  \centering
  \includegraphics[width=\textwidth]{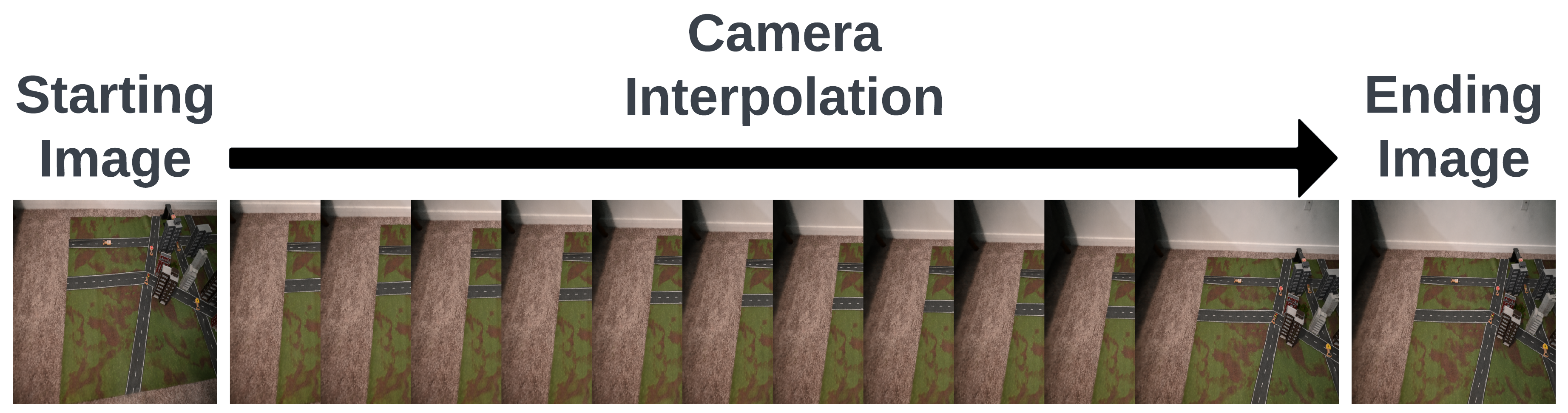}
  \caption{The provided sequence illustrates camera interpolation, where the pose transitions from the starting image to the ending image, are used to synthesize intermediate views. These interpolated views, not present in the original dataset, bridge the gap between the start and end poses, introducing new perspectives to the dataset.
}
  \label{fig:camera_interp}
\end{figure}

Given a set of non-anomalous images, an associated camera poses \( T \) can be calculated using SfM. The calculated \( T \) can be interconnected to form a trajectory. A simple approach for this interconnection is a greedy algorithm that selects a starting camera position and sequentially moves to the nearest neighboring camera position.  Specifically, for our experiments, we interpolate 12 intermediate camera poses between consecutive original poses in \( T \), thereby enhancing the dataset with interpolated views that increase the robustness of our AD model to various camera angles. An example of the camera interpolation for ToyCity can be seen in \figurename~ \ref{fig:camera_interp}.

\subsubsection{Query-Aligned Non-anomalous Views (QANV)}
Here, we propose a method to generate additional camera poses relative to the non-anomalous scene by localizing images from the query set. Our method employs hloc to estimate the camera pose for a query anomaly image. This necessitates an accurate reconstruction of the non-anomalous scene that is provided by the non-anomalous SfM model. Hloc provides various options for local feature detection, matching, and image similarity measurement. We opted for a combination of SuperPoint trained on Aachen and SuperGlue \cite{Detone2018SuperPoint:Description,Sattler2018BenchmarkingConditions}, as it demonstrated effective performance on our datasets on average. Nonetheless, the flexibility to switch to other methods is inherent in our approach, enabling adaptation to specific dataset characteristics. 

\subsection{Novel View Synthesis}
We selected Nerfacto \cite{Tancik2023Nerfstudio:Development} due to its balanced performance in training speed and quality compared to other methods, along with its user-friendly interface, which includes an interactive viewer. Although our method is agnostic to the specific NVS technique used, the aforementioned benefits made Nerfacto a compelling choice for this work. Continual advancements in this field, like the advent of 3D Gaussian Splattering\cite{Kerbl20233DRendering}, present opportunities for further enhancing rendering time and quality.

Once the new camera poses are selected, we construct our augmented datasets using NVS, as shown in \figurename~\ref{fig:nerf_augmentation_overview}. Example camera interpolated images can be seen in \figurename~\ref{fig:nerf_augmentation_overview}, and example images using QANV can be seen in \figurename~\ref{fig:nerf_augmentation_overview}

\subsection{OmniAD}
\begin{figure}
  \centering
  \includegraphics[width=0.6\textwidth]{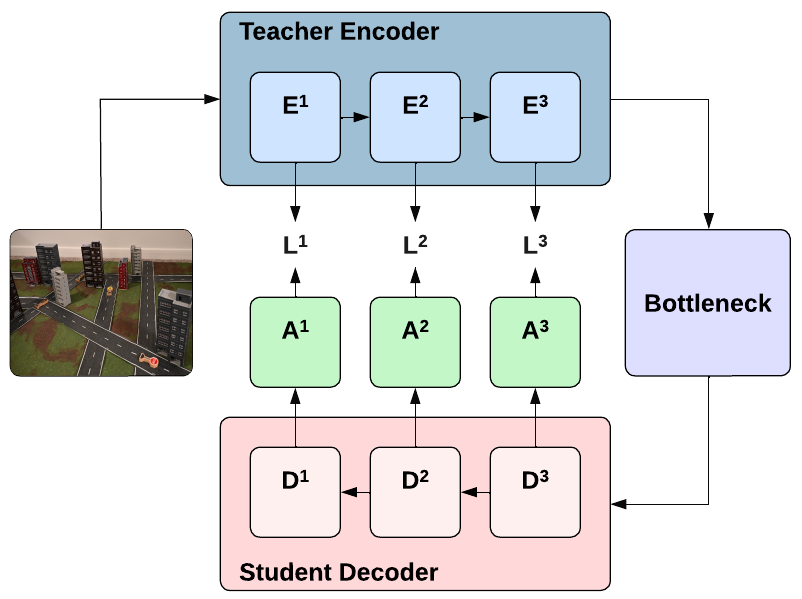}
  \caption{OmniAD's architecture mirrors RD's, featuring a frozen teacher encoder that channels features through a bottleneck module into a student decoder. We enhance this setup by integrating student attention modules \( A^1 \), \( A^2 \), and \( A^3 \).}
  \label{fig:network_arch}
\end{figure}

Our goal in Scene AD is to handle significant camera misalignment while detecting anomalies at a pixel level. Although Reverse Distillation\cite{Deng2022AnomalyEmbedding} provides a robust teacher student mechanism for fixed view scenarios, our preliminary tests revealed its limitations in multi-view scenarios. As such, OmniAD is proposed to refine Reverse Distillation to improve its generalizability in Scene AD, an overview of OmniAD is shown in \ref{fig:network_arch}.

\textbf{Backbone Exploration.}  
We first explored different backbones to replace RD's default encoder, focusing on architectures that could accommodate the diverse viewpoints. Many alternatives struggled to converge or offered minimal gains, but ResNeXt\cite{Xie2017AggregatedNetworks} consistently yielded better multiscale representations and improved performance compared to simpler encoders. In principle, any sufficiently expressive architecture including fully attention-based backbones could be substituted for the encoder. However, given the ever-growing variety of such networks, we restricted our exploration to widely adopted backbones and found that ResNeXt provided the most consistent gains for our multi-view setting.

\textbf{Receptive Field Constraints.}  
Further analysis indicated that Reverse Distillation's student decoder had a relatively small effective receptive field (ERF) \cite{Liu2018UnderstandingSegmentation}, causing it to overlook anomalies. This shortcoming proved critical in multi-view settings, where anomalies could appear in various positions or at different scales. 

\textbf{Rationale of ERF Expansion for Anomaly Sensitivity.}
Let $\mathcal{I}$ denote the image domain (e.g., $H \times W$ pixels), and let $x \in \mathbb{R}^{C \times H \times W}$ be an input image. Suppose a \textbf{teacher} network $T$ extracts multiscale features, 
\begin{equation}
  T(x) \;=\; \bigl(T^1(x), T^2(x), \dots, T^L(x)\bigr),
\end{equation}
where $T^l(x) \in \mathbb{R}^{C_l \times H_l \times W_l}$ is the $l$-th feature map. A \textbf{student} network $S$ decodes these teacher features to produce a reconstruction $\hat{x} = (S \circ T)(x)$. In RD \cite{Deng2022AnomalyEmbedding}, any mismatch between $x$ and $\hat{x}$ can indicate an anomaly, since $S$ is trained on non-anomalous data alone $S$ is unable to reconstruct anomalous features from a teacher.

\textbf{Effective Receptive Field (ERF).}
For each output location $(i, j)$ in $\hat{x}$, define its \textbf{effective receptive field} as
\begin{equation}
  \mathrm{ERF}_{ij}(S \circ T) \;=\; 
  \bigl\{\,(u, v) \in \mathcal{I} \;\big|\; 
      \tfrac{\partial \hat{x}_{ij}}{\partial x_{uv}} \neq 0
  \bigr\}.
\end{equation}
In standard RD, convolutions and limited skip connections yield a relatively localized ERF. If a true anomaly lies outside $\mathrm{ERF}_{ij}$ for many $(i, j)$, it may not sufficiently alter $\hat{x}$, causing the anomaly to go undetected.

\begin{figure}
    \centering
    \includegraphics[width=.5\linewidth]{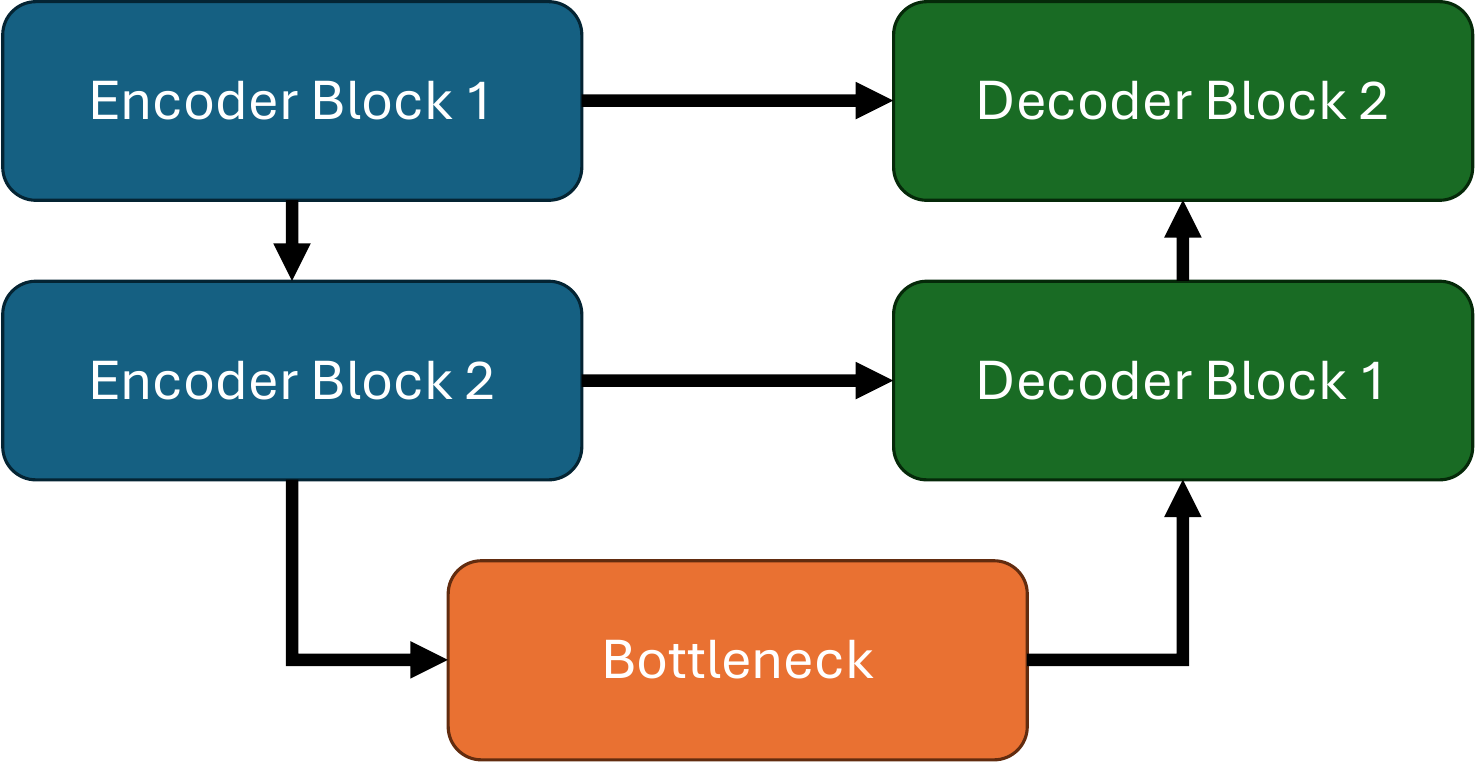}
    \caption{Student attention module layout.}
    \label{fig:student_attention_arch}
\end{figure}

\textbf{Impact of ERF Expansion.}
Let $A \subset \mathcal{I}$ denote the \textbf{anomaly region}, i.e., pixels in $x$ that deviate from normal. By adding \textbf{student attention modules} (Fig.~\ref{fig:student_attention_arch}), the new student $S'$ broadens $\mathrm{ERF}_{ij}(S' \circ T)$,
\begin{equation}
  \bigl|\mathrm{ERF}_{ij}(S' \circ T)\bigr|
  \;\;\ge\;\;
  \bigl|\mathrm{ERF}_{ij}(S \circ T)\bigr|.
\end{equation}
Consequently, for any anomalous pixel $(u, v) \in A$, there are more $(i, j)$ in the reconstructed output $\hat{x}$ for which $(u,v) \in \mathrm{ERF}_{ij}(S' \circ T)$. In other words, the anomaly influences a \textbf{broader swath} of the reconstruction, resulting in larger or more easily detectable errors. 

\textbf{Effect of ERF on Anomaly Score.}
As in conventional RD, we define an \textbf{anomaly score} per pixel $(i,j)$ by measuring the discrepancy between $\hat{x}_{ij}$ and $x_{ij}$. When the ERF is small, an anomaly's influence might be confined or even “explained away” by the network averaging out the region, yielding low detection sensitivity. With an expanded ERF, the anomaly is more likely to cause a significant mismatch for at least one output location $(i, j)$, thereby increasing detection rates.

\textbf{Expansion of ERF with Student Attention Modules.}
  
\begin{figure}
    \centering
    \subfloat[with \( A^i \)]{\includegraphics[width=1.5in, trim={0cm 1cm 1cm 0cm}, clip]{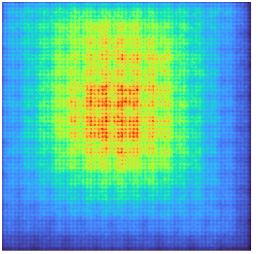}%
    \label{fig_ERF_a}}
    \hfil
    \subfloat[w/o \( A^i \)]{\includegraphics[width=1.5in, trim={0cm 1cm 1cm 0cm}, clip]{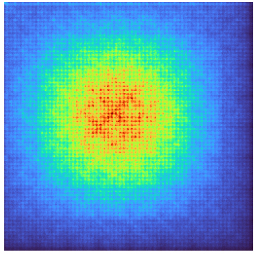}%
    \label{fig_ERF_b}}
    \caption{ERF of OmniAD with and without the student attention module. (a) with \( A^i \). (b) w/o \( A^i \).}
    \label{fig_ERF}
\end{figure}

To address the ERF limitation, we introduced additional attention layers, denoted $A^i$, after each decoder block. These modules expand the ERF, as shown in Figure~\ref{fig_ERF},  by allowing the network to aggregate information from spatially distant regions, thus capturing subtle or pose-dependent anomalies. Figure~\ref{fig:student_attention_arch} shows the basic layout of each attention module and relevant parameters are shown in Table~\ref{tab:student_attention_module_generic_simplified_dec_conv}.

\begin{table*}[htbp] 
  \centering 
  \caption{Student attention module parameters and layers:
  \textbf{\( A^3 \)}: $C_{in}=256, H_{in}=64, C_{out}=256, H_{out}=64$.
  \textbf{\( A^2 \)}: $C_{in}=512, H_{in}=32, C_{out}=512, H_{out}=32$.
  \textbf{\( A^1 \)}: $C_{in}=1024, H_{in}=16, C_{out}=1024, H_{out}=16$.
  Batch Normalization (BN), 2D Max Pooling (MaxPool2D), 2D Convolution (Conv2D)
  }
  \label{tab:student_attention_module_generic_simplified_dec_conv}
  \resizebox{\textwidth}{!}{
  \begin{tabular}{@{}llll@{}} 
  \hline
  \textbf{Stage}         & \textbf{Layer Type}                 & \textbf{Details}                                                                      & \textbf{Output Shape} \\
  \hline
  Input         & -                          & Input Channels: $C_{in}$, Spatial Size: $H_{in} \times H_{in}$                         & $B \times C_{in} \times H_{in} \times H_{in}$ \\
  \hline
  \textbf{Encoder Block 1} & Conv2D          & kernel=3, stride=1, padding=1, $C_{in} \to 64$                                       & $B \times 64 \times H_{in} \times H_{in}$ \\
                & Self Attention              & channels=64                                                                  & $B \times 64 \times H_{in} \times H_{in}$  \\
                & MaxPool2D                  & kernel=2, stride=2                                                           & $B \times 64 \times H_{in}/2 \times H_{in}/2$ \\
  \hline
  \textbf{Encoder Block 2} & Conv2D          & kernel=3, stride=1, padding=1, $64 \to 128$                                  & $B \times 128 \times H_{in}/2 \times H_{in}/2$ \\
                & Self Attention              & channels=128                                                                 & $B \times 128 \times H_{in}/2 \times H_{in}/2$  \\
                & MaxPool2D                  & kernel=2, stride=2                                                           & $B \times 128 \times H_{in}/4 \times H_{in}/4$ \\
  \hline
  \textbf{Bottleneck} & Conv2D, BN, ReLU     & kernel=3, stride=1, padding=1, $128 \to 256$                                 & $B \times 256 \times H_{in}/4 \times H_{in}/4$ \\
                & Self Attention              & channels=256                                                                 & $B \times 256 \times H_{in}/4 \times H_{in}/4$ \\
                & Conv2D, BN, ReLU     & kernel=3, stride=1, padding=1, $256 \to 256$                                 & $B \times 256 \times H_{in}/4 \times H_{in}/4$ \\
                & Self Attention              & channels=256                                                                 & $B \times 256 \times H_{in}/4 \times H_{in}/4$ \\
                & Conv2D, BN, ReLU     & kernel=3, stride=1, padding=1, $256 \to 256$                                 & $B \times 256 \times H_{in}/4 \times H_{in}/4$ \\
                & Self Attention              & channels=256                                                                 & $B \times 256 \times H_{in}/4 \times H_{in}/4$ \\
                & Conv2D, BN             & kernel=3, stride=1, padding=1, $256 \to 128$                                 & $B \times 128 \times H_{in}/4 \times H_{in}/4$ \\
  \hline
  \textbf{Decoder Block 1} & Upsample        & Target: $H_{in}/2 \times H_{in}/2$                                           & $B \times 128 \times H_{in}/2 \times H_{in}/2$ \\
                & Conv2D                     & kernel=3, stride=1, padding=1, $256 \to 64$                                  & $B \times 64 \times H_{in}/2 \times H_{in}/2$ \\
  \hline
  \textbf{Decoder Block 2} & Upsample        & Target: $H_{in} \times H_{in}$                                              & $B \times 64 \times H_{in} \times H_{in}$ \\
                & Conv2D                     & kernel=3, stride=1, padding=1, $128 \to C_{out}$                             & $B \times C_{out} \times H_{in} \times H_{in}$ \\
  \hline
  \end{tabular}%
  } 
  \end{table*}

\section{ToyCity Dataset}
\label{sec:dataset}
\begin{figure}
  \centering
  \includegraphics[width=\textwidth]{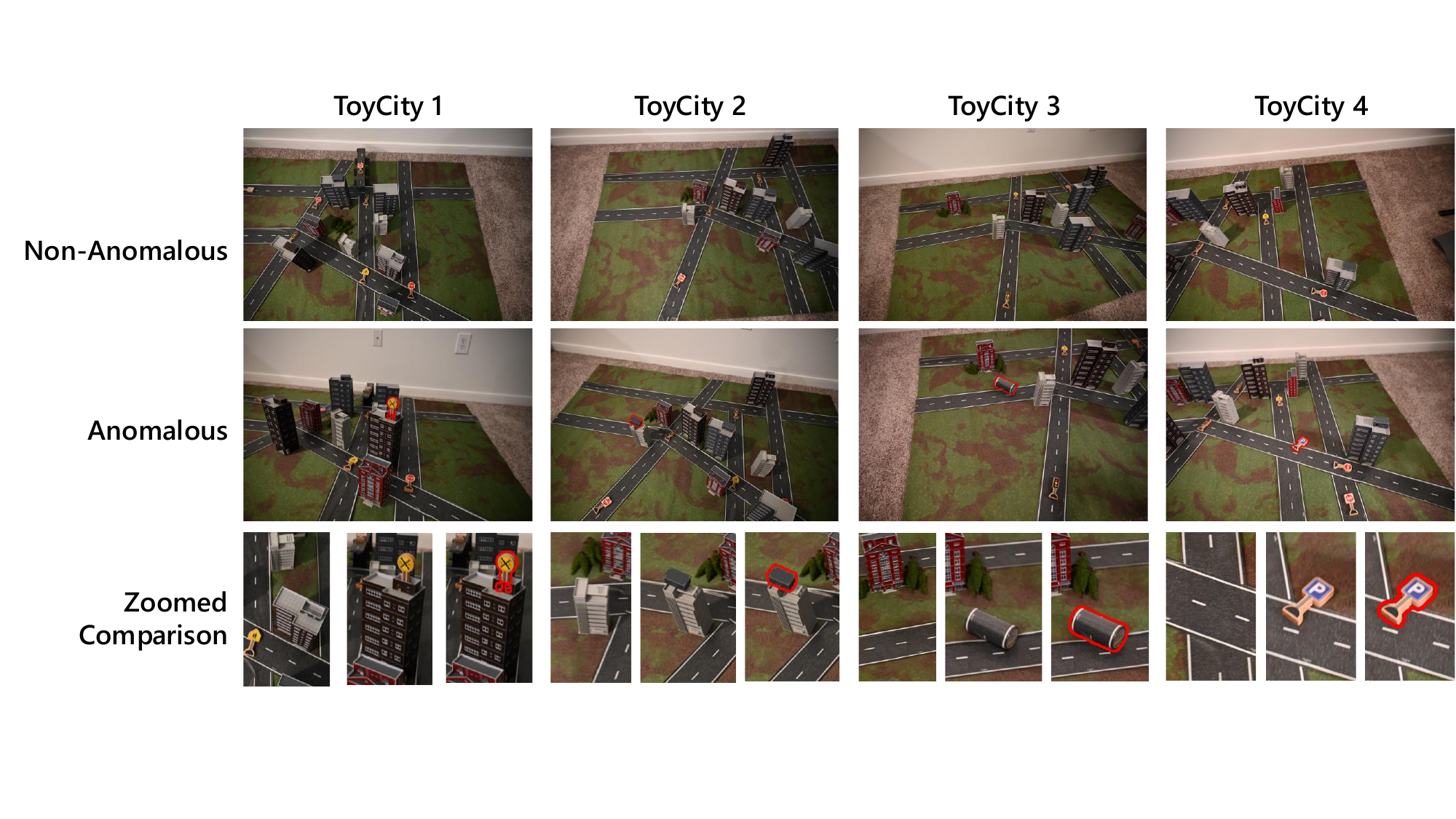}
  \caption{ToyCity consists of four configurations as shown here. We present an example baseline image for each, as well as an example of one of several anomalies used. A zoomed comparison is also presented that shows the anomaly highlighted with a red outline.}
  \label{fig:dataset_anomaly}
\end{figure}

\begin{figure}
  \centering
  \includegraphics[height=0.65\textwidth, keepaspectratio]{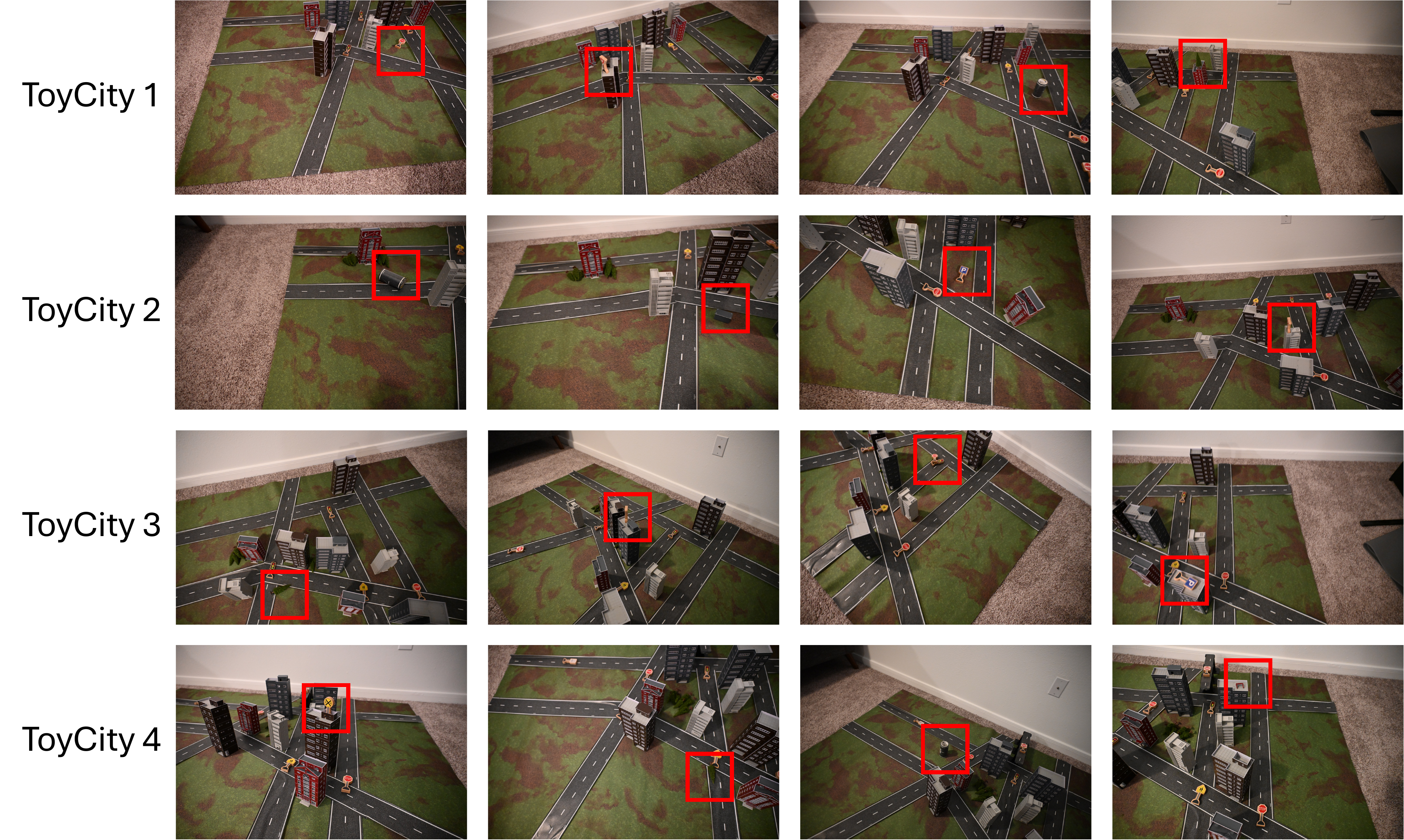}
  \caption{Each ToyCity configuration has several anomalies that have been added and removed from the scene. Examples of these anomalies for each ToyCity are shown here by a red bounding box.}
  \label{fig:dataset_additional_anomalies_landscape} 
\end{figure}

The ToyCity dataset addresses the need for a real image benchmark dataset consisting of multiple objects and multiple views. As shown in \tablename~\ref{tab:existing_anomaly_dataset_comparisons_modified}, ToyCity is comparable in size to existing prominent AD benchmarks that consist of real images of anomalies. Additionally, ToyCity consists of substantially more real images, consists of unconstrained random views, and is not centered around a single object. ToyCity features scenes of a toy city with buildings, signs, roads, trees, and miscellaneous items. The dataset contains images from various angles and heights in four distinct scene arrangements, as illustrated in \figurename~\ref{fig:dataset_anomaly}. Each setup also contains different anomalies, as shown in \figurename~\ref{fig:dataset_additional_anomalies_landscape}. Additionally, each anomaly image has at most one anomaly, but each setup has varying anomalies that were added and removed as shown in \figurename~\ref{fig:dataset_additional_anomalies_landscape}. Anomalous images contain on average $0.22\%$ anomalous pixels (standard deviation $0.13\%$). The percentage of anomalous pixels per image ranges from a minimum of $0.03\%$ to a maximum of $0.78\%$. Table \ref{tab:dataset_detailed_variation_count} provides an overview of the number of images released for ToyCity. Each ToyCity has four augmented variations, without any augmentations, with Query-Aligned Novel Views (QANV), with Interpolated Non-Anomalous Views (INV), and combining both QANV and INV; these variations are presented in Table \ref{tab:dataset_detailed_variations}. ToyCity thus offers 16 total variations that will be evaluated.

\begin{table}[htbp]
  \centering 
  \caption{Comparison of datasets containing real images of anomalies with ground truth labels for quantitative evaluation.} 
  \label{tab:existing_anomaly_dataset_comparisons_modified} 
  \resizebox{\textwidth}{!}{
  \begin{tabular}{@{}lcccc@{}} 
  \hline
  \textbf{Dataset} & \textbf{Data Availability} & \textbf{Labeled Anomalous Images} & \textbf{Multi-View} & \textbf{Multi-Object} \\
  \hline
  AITEX\cite{silvestre2019public} & Yes & 105 & No & No \\
  MPDD\cite{jezek2021deep} & Yes & 282 & No & Yes \\
  MVTec-AD\cite{bergmann2019mvtec} & Yes & 1258 & No & No \\
  MVTec 3D-AD\cite{bergmann2021mvtec} & Yes & 948 & No & No \\
  MVTec LOCO-AD\cite{bergmann2022beyond} & Yes & 993 & No & Yes \\
  NanoTwice\cite{carrera2016defect} & Yes & 40 & No & No \\
  VisA\cite{zou2022spot} & Yes & 1200 & No & Yes \\
  MAD-Real\cite{ZhouPad:Detection}& Partial & 221 & Yes & No\\
  \hline
  ToyCity& Yes & 1258 & Yes & Yes\\
  \hline 
  \end{tabular}
  }
\end{table}

\begin{table}
\centering
\caption{Number of non-anomalous and anomalous images in each of the ToyCity dataset scenes, and the number of NVS images generated from each of the proposed methods QANV and IBV}
\label{tab:dataset_detailed_variation_count}
\setlength\tabcolsep{2pt}
\begin{tabular}{c c c c c c c}
\hline
\textbf{Name} & \textbf{Non-Anomalous} & \textbf{Anomalous} & \textbf{QANV} & \textbf{IBV} & \textbf{Images}\\
\hline
\textbf{ToyCity 1} & 316 & 319 & 319 & 3781 & 4735\\
\hline
\textbf{ToyCity 2} & 341 & 308 & 308 & 3773 & 4730\\
\hline
\textbf{ToyCity 3} & 307 & 317 & 317 & 3673 & 4614\\
 \hline
\textbf{ToyCity 4} & 314 & 314 & 314 & 3757 & 4699\\
\hline
\end{tabular}
\end{table}

\def\checkmark{\tikz\fill[scale=0.4](0,.35) -- (.25,0) -- (1,.7) -- (.25,.15) -- cycle;} 

\section{Experiments}
\label{sec:experiments}

We evaluate OmniAD's performance, its ablated variants, and several AD methods on ToyCity and MAD to test the localization ability of each method. As the camera poses used in MAD-Real \cite{ZhouPad:Detection} for testing were not released publicly, we could only utilize a subset of the MAD dataset that COLMAP could generate camera poses for. Each AD method is evaluated quantitatively using the pixel-level \( F_1 \) score and AUROC. Additionally, we evaluate the performance of OmniAD on the MVTec-AD dataset \cite{bergmann2019mvtec}. An ablation study follows to show the significance of the student attention modules on Reverse Distillation. Finally, a qualitative comparison between OmniAD and Reverse Distillation for anomaly segmentation is presented.

\subsection{Metrics}
\label{sec:metrics}
To comprehensively assess the anomaly localization performance of various AD models, we employ the pixel-wise \( F_1 \), 
\begin{equation} \label{eq:F1_score}
  F_1 = 2 \cdot \frac{\text{precision} \cdot \text{recall}}{\text{precision} + \text{recall}}.
\end{equation} The \( F_1 \) score excels in handling class imbalances by merging precision and recall into one comprehensive metric.
This balance is critical for datasets such as ToyCity and MAD, where anomaly regions typically constitute a small fraction of the total image area. Examples of the relatively small anomaly regions within ToyCity can be seen in \figurename~ \ref{fig:dataset_anomaly}, underscoring the necessity for such a balanced evaluation metric. The imbalance of anomaly regions renders other metrics like accuracy or the area under the receiver operating characteristic curve (AUROC) as misleading.

The reason accuracy is deceptive is that a high accuracy metric can be achieved with a simplistic approach that labels all pixels as non-anomalous. This approach would yield numerous true negatives and few false positives. A similar effect can be observed with the AUROC, which may show a more optimistic estimate of model performance. However, it does not accurately reflect the model's ability to precisely localize anomalies, primarily due to the abundance of true negatives. Thus, we provide both the \( F_1 \) score and AUROC.

\subsection{Implementation and Training Details}
To ensure a fair comparison and promote reproducibility, all anomaly detection methods evaluated in this study (including CFLOW-AD, FastFlow, CFA, DRÆM, RD, and our proposed OmniAD variants) were implemented using the publicly available anomalib library \cite{anomalib}, which has faithfully integrated each method based on original implementations and allows for reproducibility.

\textbf{Parameter Settings:} For all baseline methods, we utilized the default hyperparameter configurations provided within the anomalib library to match original implementations. OmniAD and its ablated versions also inherit these same default hyperparameters from the anomalib implementation of the underlying Reverse Distillation method. This approach ensures that performance differences observed are primarily attributable to the architectural modifications (i.e., the ResNeXt backbone and the student attention modules in OmniAD) and the novel view synthesis data augmentation strategies, rather than method-specific hyperparameter tuning.

\textbf{Training Procedure:} All models were trained for a maximum of 100 epochs, using a batch size of 32 for fair comparison. Early stopping was implicitly handled by selecting the model checkpoint that yielded the maximum pixel-level \( F_1 \) score on a validation set, consistent with standard practices. We observed that all methods converged well before the 100-epoch limit, ensuring adequate training time. Learning rates and model specific training hyperparameters follow the anomalib defaults for each model.

\subsection{Quantitative Evaluation}

In our assessment of diverse AD methods for Scene AD using ToyCity and MAD-Real\cite{ZhouPad:Detection}, \tablename~\ref{tab:toycity_1_2},\ref{tab:toycity_3_4},\ref{tab:f1_mad_combined}, we observe a consistent performance increase for OmniAD compared to Reverse Distillation. A similar trend can also be seen for AUROC \tablename~\ref{tab:auroc_toycity_1_2},\ref{tab:auroc_toycity_3_4},\ref{tab:auroc_mad_combined}.

The Reverse Distillation with ResNext model and the canonical Reverse Distillation methodology exhibit similar enhancement trends, yet they exhibit smaller gains in comparison to our model. The shift from no augmentation to QANV, where query aligned images are introduced, is particularly telling of the significant enhancements in the models' ability to localize anomalies due to augmentation. Other AD methods such as CFLOW, FastFlow, CFA, and DRÆM show a range of responses to the augmentations, with \( F_1 \) scores experiencing minimal improvements, remaining unchanged, or occasionally declining with the introduction of more complex augmentations. This suggests a potential lack of effective utilization of augmented data by these methods in contrast to OmniAD and Reverse Distillation.

As mentioned, metrics such as pixel-level AUROC can often show very high numbers for datasets, such as ToyCity and MAD, due to the low number of pixels that may be an anomaly. This can be seen for both ToyCity \tablename~\ref{tab:auroc_toycity_1_2},\ref{tab:auroc_toycity_3_4} and MAD \tablename~\ref{tab:auroc_mad_combined}. Most AD methods will achieve an AUROC very close to 1 while still offering low \( F_1 \) scores.

To evaluate the generalizability of OmniAD beyond Scene AD, we also test its performance in a standard fixed-view AD setting using the MVTec-AD dataset\cite{bergmann2019mvtec}. As indicated in \tablename~\ref{tab:sota_mvtec_ad}, OmniAD remains effective in this fixed-view context. Although the high AUROC (0.975) achieved by Reverse Distillation for fixed-view AD limits newer methods like OmniAD to incremental improvements until a perfect AUROC is attained. OmniAD does demonstrate a more substantial performance gain in terms of the \( F_1\) score compared to Reverse Distillation and other recent methods such as SimpleNet\cite{liu2023simplenet}, MambaAD\cite{He2024MambaADES}, and AnomalyDINO\cite{damm2025anomalydino}.


\begin{table}
\centering
\caption{Four augmentation variations are tested for each of the scenes present in ToyCity.}
\label{tab:dataset_detailed_variations}
\setlength\tabcolsep{2pt}
\begin{tabular}{c c c c c c c}
\hline
\textbf{Variation} & \textbf{Non-Anomalous} & \textbf{QANV} & \textbf{INV} & \textbf{Images} \\
\hline

\textbf{No Augmentation} & \checkmark & - & - & 300+\\
\hline
\textbf{QANV Augmentation} & \checkmark  & \checkmark & - & 600+\\
\hline
\textbf{INV Augmentation} & \checkmark  & - & \checkmark& 3600+\\
\hline
\textbf{Both QANV and INV} & \checkmark  & \checkmark & \checkmark& 3900+\\
\hline
\end{tabular}
\end{table}

\subsection{Qualitative Evaluation}
\begin{figure*}
  \centering
  \includegraphics[width=\textwidth]{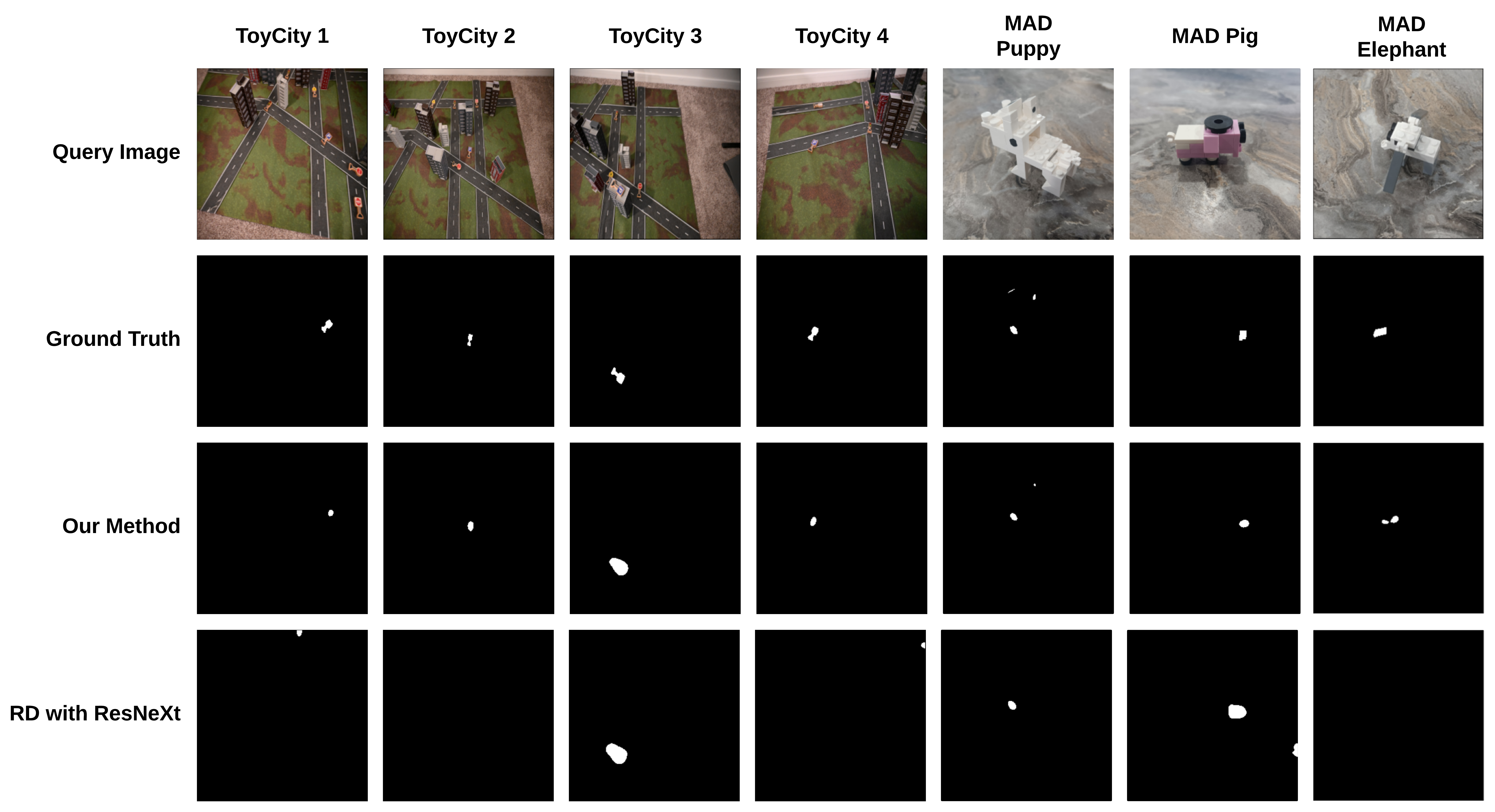}
  \caption{Segmentation performance comparison between the top two evaluated AD methods OmniAD and Reverse Distillation with a ResNeXt backbone across four ToyCity anomalies and objects from MAD. We illustrate three primary failure patterns: misidentifying of non-anomalous areas (ToyCity 1,4, and MAD Pig), overlooking anomalies (ToyCity 2 and Mad Puppy), and overestimation of anomaly areas (ToyCity 3 and MAD Pig). Other evaluated methods also fail in the same manner as Reverse Distillation  with a ResNeXt, but at a higher frequency.}
  \label{fig:segmentation}
\end{figure*}

\begin{figure*}
  \centering
  \includegraphics[width=\textwidth]{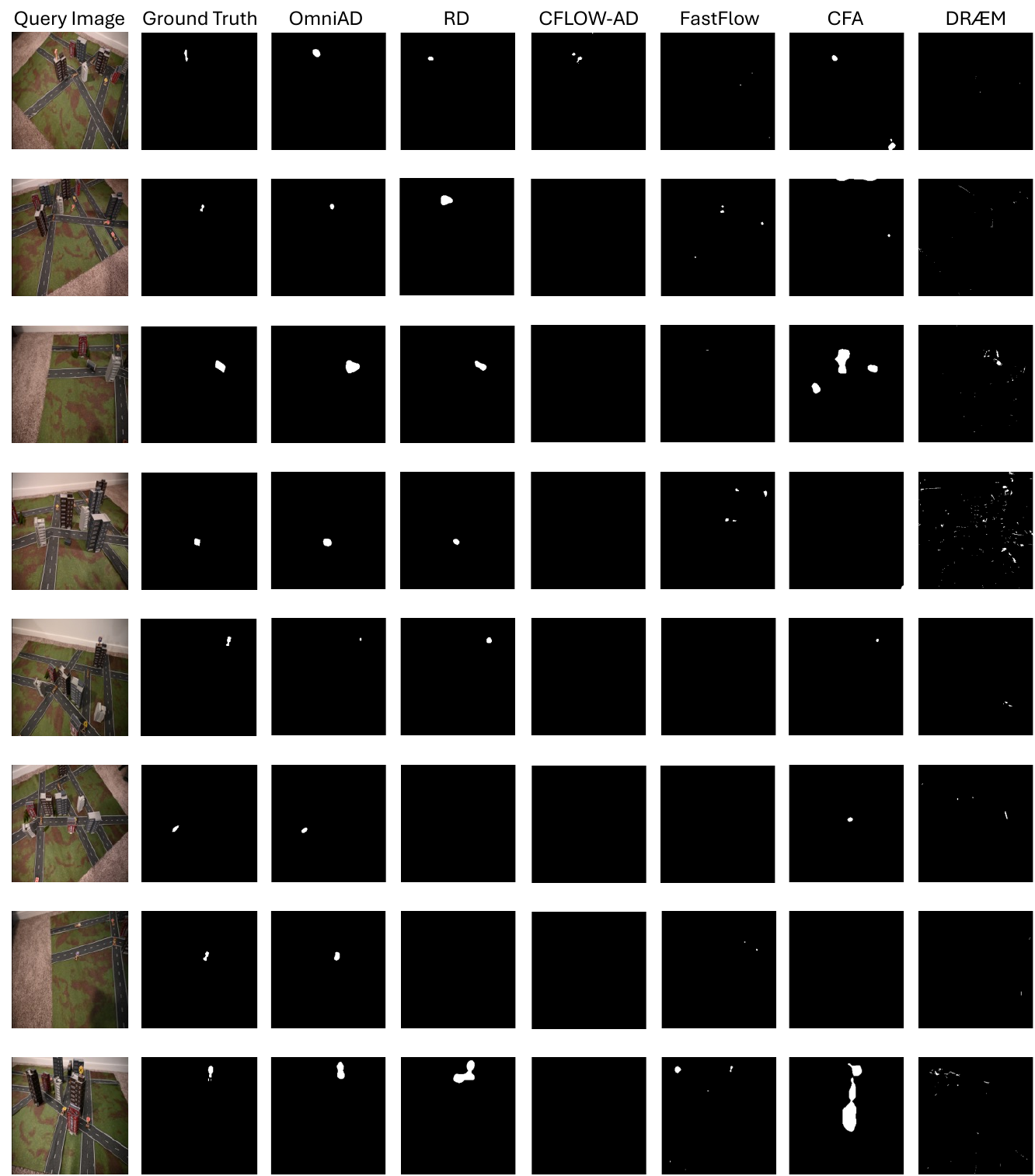}
  \caption{Additional segmentations across all evaluated AD methods.}
  \label{fig:Additional_Results_All_Methods}
\end{figure*}

\figurename~\ref{fig:segmentation} illustrates the comparative segmentation capabilities of OmniAD and the second-best performing method, Reverse Distillation, which utilizes a ResNeXt backbone. A threshold was selected for each method that maximizes their pixel-level \( F_1 \) score. Three primary failure scenarios are evident. First, as depicted in the ToyCity 1 and 4 anomaly instances, Reverse Distillation may erroneously identify arbitrary areas as anomalous. Secondly, Reverse Distillation has a propensity to overlook entire anomalous regions, a shortcoming exemplified by the ToyCity 2 anomaly. Lastly, Reverse Distillation often overestimates the extent of an anomaly, erroneously marking larger areas than necessary. Similar defects in Reverse Distillation can also be seen from the MAD dataset, in which it misses an anomaly or overestimates an anomalous region. Additional samples from OmniAD and Reverse Distillation with ResNeXt for each dataset are presented in Figure \ref{fig:Additional_Results}, showing a heat map of the anomaly score as well as the segmentation. Additional comparisons between evaluated methods across different anomalies are shown in \figurename~\ref{fig:Additional_Results_All_Methods}.

\begin{figure*}
  \centering
  \includegraphics[width=\textwidth]{OmniAD_Results.pdf}
  \caption{Additional segmentation and anomaly scores from OmniAD.}
  \label{fig:Additional_Results}
\end{figure*}

\begin{table}
    \centering
    \caption{Comparison of Pixel \( F_1 \) scores on ToyCity 1 and 2}
    \label{tab:toycity_1_2}
    \setlength\tabcolsep{4pt}
    \resizebox{\linewidth}{!}{
    \begin{tabular}{lcccc|cccc}
    \hline
    \multirow{2}{*}{\textbf{Method}} & \multicolumn{4}{c|}{\textbf{ToyCity 1}} & \multicolumn{4}{c}{\textbf{ToyCity 2}} \\
    \cline{2-9}
     & \textbf{No Aug} & \textbf{QANV} & \textbf{INV} & \textbf{Both} & \textbf{No Aug} & \textbf{QANV} & \textbf{INV} & \textbf{Both} \\
    \hline
    \textbf{OmniAD} & 0.432 & 0.450 & \textbf{0.497} & 0.496 & 0.348 & \textbf{0.599} & 0.491 & 0.577 \\
    \textbf{OmniAD w/o R} & 0.422 & 0.448 & 0.431 & 0.445 & 0.363 & 0.531 & 0.383 & 0.503 \\
    \textbf{OmniAD w/o $A^i$} & 0.391 & 0.411 & 0.409 & 0.388 & 0.256 & 0.546 & 0.476 & 0.357 \\
    \textbf{RD} \cite{Deng2022AnomalyEmbedding} & 0.307 & 0.364 & 0.365 & 0.327 & 0.317 & 0.350 & 0.388 & 0.390 \\
    \textbf{CFLOW-AD} \cite{Gudovskiy2022CFLOW-AD:Flows} & 0.190 & 0.218 & 0.088 & 0.052 & 0.147 & 0.183 & 0.105 & 0.082 \\
    \textbf{FastFlow} \cite{Yu2021FastFlow:Flows} & 0.124 & 0.135 & 0.080 & 0.079 & 0.173 & 0.159 & 0.094 & 0.128 \\
    \textbf{CFA} \cite{Lee2022Cfa:Localization} & 0.109 & 0.083 & 0.072 & 0.078 & 0.092 & 0.090 & 0.056 & 0.081 \\
    \textbf{DRÆM} \cite{Zavrtanik2021DRMDetection} & 0.031 & 0.018 & 0.043 & 0.047 & 0.034 & 0.092 & 0.041 & 0.039 \\
    \hline
    \end{tabular}
    }
    \end{table}
    
    \begin{table}
    \centering
    \caption{Comparison of Pixel \( F_1 \) scores on ToyCity 3 and 4}
    \label{tab:toycity_3_4}
    \setlength\tabcolsep{4pt}
    \resizebox{\linewidth}{!}{
    \begin{tabular}{lcccc|cccc}
    \hline
    \multirow{2}{*}{\textbf{Method}} & \multicolumn{4}{c|}{\textbf{ToyCity 3}} & \multicolumn{4}{c}{\textbf{ToyCity 4}} \\
    \cline{2-9}
     & \textbf{No Aug} & \textbf{QANV} & \textbf{INV} & \textbf{Both} & \textbf{No Aug} & \textbf{QANV} & \textbf{INV} & \textbf{Both} \\
    \hline
    \textbf{OmniAD} & 0.281 & \textbf{0.384} & 0.319 & 0.366 & 0.379 & \textbf{0.541} & 0.390 & 0.523 \\
    \textbf{OmniAD w/o R} & 0.281 & 0.338 & 0.294 & 0.369 & 0.337 & 0.466 & 0.376 & 0.502 \\
    \textbf{OmniAD w/o $A^i$} & 0.196 & 0.276 & 0.243 & 0.290 & 0.279 & 0.434 & 0.284 & 0.403 \\
    \textbf{RD} \cite{Deng2022AnomalyEmbedding} & 0.112 & 0.167 & 0.171 & 0.220 & 0.270 & 0.380 & 0.257 & 0.352 \\
    \textbf{CFLOW-AD} \cite{Gudovskiy2022CFLOW-AD:Flows} & 0.066 & 0.052 & 0.056 & 0.064 & 0.093 & 0.083 & 0.132 & 0.128 \\
    \textbf{FastFlow} \cite{Yu2021FastFlow:Flows} & 0.214 & 0.210 & 0.112 & 0.111 & 0.189 & 0.188 & 0.123 & 0.112 \\
    \textbf{CFA} \cite{Lee2022Cfa:Localization} & 0.107 & 0.081 & 0.067 & 0.131 & 0.094 & 0.120 & 0.054 & 0.062 \\
    \textbf{DRÆM} \cite{Zavrtanik2021DRMDetection} & 0.029 & 0.032 & 0.044 & 0.018 & 0.047 & 0.087 & 0.064 & 0.091 \\
    \hline
    \end{tabular}
    }
    \end{table}
    
    \begin{table}
    \centering
    \caption{Comparison of Pixel AUROC scores on ToyCity 1 and 2}
    \label{tab:auroc_toycity_1_2}
    \setlength\tabcolsep{4pt}
    \resizebox{\linewidth}{!}{
    \begin{tabular}{lcccc|cccc}
    \hline
    \multirow{2}{*}{\textbf{Method}} & \multicolumn{4}{c|}{\textbf{ToyCity 1}} & \multicolumn{4}{c}{\textbf{ToyCity 2}} \\
    \cline{2-9}
     & \textbf{No Aug} & \textbf{QANV} & \textbf{INV} & \textbf{Both} & \textbf{No Aug} & \textbf{QANV} & \textbf{INV} & \textbf{Both} \\
    \hline
    \textbf{OmniAD}         & 0.990 & 0.996 & \textbf{0.999} & \textbf{0.999} & 0.991 & 0.998 & \textbf{0.999} & \textbf{0.999} \\
    \textbf{OmniAD w/o R}       & 0.988 & 0.993 & \textbf{0.999} & \textbf{0.999} & 0.991 & 0.995 & 0.997       & \textbf{0.999} \\
    \textbf{OmniAD w/o $A^i$}   & 0.988 & 0.996 & \textbf{0.999} & 0.998       & 0.988 & 0.998 & \textbf{0.999} & 0.997       \\
    \textbf{RD} \cite{Deng2022AnomalyEmbedding}             & 0.987 & 0.992 & 0.998       & 0.998       & 0.986 & 0.990 & 0.998       & 0.998       \\
    \textbf{CFLOW-AD} \cite{Gudovskiy2022CFLOW-AD:Flows}          & 0.920 & 0.914 & 0.864       & 0.921       & 0.928 & 0.934 & 0.935       & 0.868       \\
    \textbf{FastFlow} \cite{Yu2021FastFlow:Flows}       & 0.912 & 0.939 & 0.940       & 0.956       & 0.948 & 0.952 & 0.957       & 0.952       \\
    \textbf{CFA} \cite{Lee2022Cfa:Localization}            & 0.943 & 0.946 & 0.976       & 0.976       & 0.953 & 0.962 & 0.977       & 0.984       \\
    \textbf{DRÆM} \cite{Zavrtanik2021DRMDetection}           & 0.578 & 0.449 & 0.585       & 0.597       & 0.621 & 0.600 & 0.700       & 0.725       \\
    \hline
    \end{tabular}
    }
    \end{table}
    
    \begin{table}
    \centering
    \caption{Comparison of Pixel AUROC scores on ToyCity 3 and 4}
    \label{tab:auroc_toycity_3_4}
    \setlength\tabcolsep{4pt}
    \resizebox{\linewidth}{!}{
    \begin{tabular}{lcccc|cccc}
    \hline
    \multirow{2}{*}{\textbf{Method}} & \multicolumn{4}{c|}{\textbf{ToyCity 3}} & \multicolumn{4}{c}{\textbf{ToyCity 4}} \\
    \cline{2-9}
     & \textbf{No Aug} & \textbf{QANV} & \textbf{INV} & \textbf{Both} & \textbf{No Aug} & \textbf{QANV} & \textbf{INV} & \textbf{Both} \\
    \hline
    \textbf{OmniAD}         & 0.979 & 0.993 & 0.997       & \textbf{0.999} & 0.982 & 0.993 & 0.995       & \textbf{0.998} \\
    \textbf{OmniAD w/o R}       & 0.980 & 0.985 & 0.998       & 0.998       & 0.980 & 0.989 & 0.996       & \textbf{0.998} \\
    \textbf{OmniAD w/o $A^i$}   & 0.967 & 0.990 & 0.997       & 0.997       & 0.975 & 0.990 & 0.995       & 0.997       \\
    \textbf{RD} \cite{Deng2022AnomalyEmbedding}             & 0.961 & 0.987 & 0.995       & 0.996       & 0.977 & 0.990 & 0.995       & 0.997       \\
    \textbf{CFLOW-AD} \cite{Gudovskiy2022CFLOW-AD:Flows}          & 0.914 & 0.844 & 0.906       & 0.914       & 0.932 & 0.930 & 0.899       & 0.916       \\
    \textbf{FastFlow} \cite{Yu2021FastFlow:Flows}       & 0.946 & 0.932 & 0.947       & 0.936       & 0.949 & 0.947 & 0.971       & 0.972       \\
    \textbf{CFA} \cite{Lee2022Cfa:Localization}            & 0.928 & 0.953 & 0.975       & 0.979       & 0.928 & 0.953 & 0.975       & 0.979       \\
    \textbf{DRÆM} \cite{Zavrtanik2021DRMDetection}           & 0.616 & 0.628 & 0.612       & 0.626       & 0.627 & 0.570 & 0.594       & 0.626       \\
    \hline
    \end{tabular}
    }
    \end{table}

    \begin{table}
    \centering
    \caption{Comparison of Pixel \( F_1 \) scores on MAD-Real \cite{ZhouPad:Detection} of Puppy, Pig, and Elephant}
    \label{tab:f1_mad_combined}
    \setlength\tabcolsep{4pt}
    \resizebox{\linewidth}{!}{
    \begin{tabular}{lcccc|cccc|cccc}
    \hline
    \multirow{2}{*}{\textbf{Method}} & \multicolumn{4}{c|}{\textbf{Puppy}} & \multicolumn{4}{c|}{\textbf{Pig}} & \multicolumn{4}{c}{\textbf{Elephant}} \\
    \cline{2-13}
     & \textbf{No Aug} & \textbf{QANV} & \textbf{INV} & \textbf{Both} & \textbf{No Aug} & \textbf{QANV} & \textbf{INV} & \textbf{Both} & \textbf{No Aug} & \textbf{QANV} & \textbf{INV} & \textbf{Both} \\
    \hline
    \textbf{OmniAD} & 0.672 & \textbf{0.714} & 0.680 & 0.635 & 0.334 & 0.395 & 0.471 & \textbf{0.524} & 0.341 & \textbf{0.368} & 0.374 & 0.328 \\
    \textbf{OmniAD w/o R} & 0.613 & 0.677 & 0.621 & 0.623 & 0.302 & 0.346 & 0.454 & 0.441 & 0.312 & 0.204 & 0.297 & 0.284 \\
    \textbf{OmniAD w/o $A^i$} & 0.634 & 0.646 & 0.377 & 0.363 & 0.306 & 0.286 & 0.343 & 0.254 & 0.275 & 0.298 & 0.248 & 0.173 \\
    \textbf{RD} \cite{Deng2022AnomalyEmbedding} & 0.616 & 0.628 & 0.520 & 0.199 & 0.219 & 0.203 & 0.302 & 0.260 & 0.262 & 0.236 & 0.170 & 0.192 \\
    \textbf{CFLOW-AD} \cite{Gudovskiy2022CFLOW-AD:Flows} & 0.496 & 0.466 & 0.484 & 0.579 & 0.162 & 0.142 & 0.150 & 0.147 & 0.309 & 0.163 & 0.214 & 0.209 \\
    \textbf{FastFlow} \cite{Yu2021FastFlow:Flows} & 0.646 & 0.651 & 0.575 & 0.598 & 0.066 & 0.087 & 0.043 & 0.058 & 0.297 & 0.285 & 0.282 & 0.297 \\
    \textbf{CFA} \cite{Lee2022Cfa:Localization} & 0.568 & 0.576 & 0.594 & 0.067 & 0.294 & 0.232 & 0.048 & 0.029 & 0.189 & 0.151 & 0.117 & 0.139 \\
    \textbf{DRÆM} \cite{Zavrtanik2021DRMDetection} & 0.042 & 0.016 & 0.022 & 0.018 & 0.040 & 0.041 & 0.072 & 0.110 & 0.020 & 0.041 & 0.037 & 0.028 \\
    \hline
    \end{tabular}
    }
    \end{table}
    
    \begin{table}
    \centering
    \caption{Comparison of Pixel AUROC scores on MAD-Real \cite{ZhouPad:Detection} of Puppy, Pig, and Elephant}
    \label{tab:auroc_mad_combined}
    \setlength\tabcolsep{4pt}
    \resizebox{\linewidth}{!}{
    \begin{tabular}{lcccc|cccc|cccc}
    \hline
    \multirow{2}{*}{\textbf{Method}} & \multicolumn{4}{c|}{\textbf{Puppy}} & \multicolumn{4}{c|}{\textbf{Pig}} & \multicolumn{4}{c}{\textbf{Elephant}} \\
    \cline{2-13}
     & \textbf{No Aug} & \textbf{QANV} & \textbf{INV} & \textbf{Both} & \textbf{No Aug} & \textbf{QANV} & \textbf{INV} & \textbf{Both} & \textbf{No Aug} & \textbf{QANV} & \textbf{INV} & \textbf{Both} \\
    \hline
    \textbf{OmniAD} & \textbf{0.999} & \textbf{0.999} & \textbf{0.999} & \textbf{0.999} & 0.988 & 0.993 & \textbf{0.999} & \textbf{0.999} & 0.941 & 0.968 & \textbf{0.973} & 0.971 \\
    \textbf{OmniAD w/o R} & \textbf{0.999} & 0.993 & \textbf{0.999} & \textbf{0.999} & 0.988 & 0.983 & \textbf{0.999} & \textbf{0.999} & 0.924 & 0.892 & 0.933 & 0.928 \\
    \textbf{OmniAD w/o $A^i$} & \textbf{0.999} & \textbf{0.999} & 0.998 & \textbf{0.999} & 0.983 & 0.988 & 0.997 & 0.997 & 0.948 & 0.946 & 0.966 & 0.963 \\
    \textbf{RD} \cite{Deng2022AnomalyEmbedding} & \textbf{0.999} & \textbf{0.999} & \textbf{0.999} & \textbf{0.999} & 0.981 & 0.982 & 0.998 & 0.998 & 0.936 & 0.929 & 0.951 & 0.956 \\
    \textbf{CFLOW-AD} \cite{Gudovskiy2022CFLOW-AD:Flows} & 0.996 & 0.996 & \textbf{0.999} & \textbf{0.999} & 0.962 & 0.967 & 0.993 & 0.990 & 0.945 & 0.906 & 0.947 & 0.953 \\
    \textbf{FastFlow} \cite{Yu2021FastFlow:Flows} & \textbf{0.999} & 0.998 & \textbf{0.999} & \textbf{0.999} & 0.868 & 0.896 & 0.965 & 0.966 & 0.862 & 0.914 & 0.934 & 0.932 \\
    \textbf{CFA} \cite{Lee2022Cfa:Localization} & 0.998 & 0.998 & \textbf{0.999} & 0.994 & 0.977 & 0.976 & 0.986 & 0.979 & 0.916 & 0.927 & 0.934 & 0.942 \\
    \textbf{DRÆM} \cite{Zavrtanik2021DRMDetection} & 0.819 & 0.359 & 0.514 & 0.524 & 0.720 & 0.571 & 0.684 & 0.796 & 0.601 & 0.496 & 0.559 & 0.519 \\
    \hline
    \end{tabular}
    }
    \end{table}

\begin{table*} 
\caption{Comparison of pixel wise AUROC/\( F_1 \) metrics on the MVTec-AD\cite{bergmann2019mvtec} dataset.}
\label{tab:sota_mvtec_ad}
\centering
\resizebox{\textwidth}{!}{%
\begin{tabular}{@{}llcccccc@{}} 
\hline
\multicolumn{2}{@{}l}{\textbf{Method}} & OmniAD & RD\cite{Deng2022AnomalyEmbedding} & UniAD\cite{you2022unified} & SimpleNet\cite{liu2023simplenet} & AnomalyDINO\cite{damm2025anomalydino}* & MambaAD\cite{He2024MambaADES}\\
\hline
\multirow{10}{*}{\rotatebox[origin=c]{90}{Objects\hspace*{1em}}}
& Bottle & 0.987/0.755 & 0.986/0.741 & 0.983/0.707 & 0.980/0.727 & -- & \textbf{0.988}/\textbf{0.766} \\
& Cable & \textbf{0.983}/\textbf{0.633} & 0.968/0.573 & 0.974/0.569 & 0.974/0.600 & -- & 0.980/0.617 \\
& Capsule & 0.984/\textbf{0.509} & \textbf{0.989}/0.499 & 0.980/0.401 & \textbf{0.989}/0.491 & -- & 0.986/0.475 \\
& Hazelnut & \textbf{0.993}/\textbf{0.710} & 0.988/0.608 & 0.983/0.561 & 0.979/0.503 & -- & 0.990/0.639 \\
& Metal Nut & 0.987/0.865 & 0.969/0.795 & 0.944/0.688 & \textbf{0.988}/\textbf{0.866} & -- & 0.971/0.806 \\
& Pill & 0.983/0.703 & 0.976/0.667 & 0.952/0.519 & \textbf{0.985}/\textbf{0.726} & -- & 0.973/0.660 \\
& Screw & 0.993/0.479 & \textbf{0.995}/0.472 & 0.984/0.236 & 0.993/0.388 & -- & \textbf{0.995}/\textbf{0.517} \\
& Toothbrush & \textbf{0.992}/\textbf{0.627} & 0.990/0.609 & 0.985/0.521 & 0.985/0.529 & -- & 0.990/0.607 \\
& Transistor & 0.967/0.632 & 0.906/0.549 & \textbf{0.988}/0.585 & 0.968/0.621 & -- & 0.968/\textbf{0.675} \\
& Zipper & 0.988/\textbf{0.686} & 0.987/0.590 & 0.965/0.434 & \textbf{0.989}/0.654 & -- & 0.981/0.588 \\
\hline
\multirow{5}{*}{\rotatebox[origin=c]{90}{Textures\hspace*{1em}}}
& Carpet & 0.986/\textbf{0.666} & 0.991/0.609 & 0.985/0.533 & 0.980/0.478 & -- & \textbf{0.992}/0.634 \\
& Grid & \textbf{0.994}/\textbf{0.523} & 0.993/0.477 & 0.943/0.315 & 0.988/0.393 & -- & 0.992/0.488 \\
& Leather & \textbf{0.995}/\textbf{0.575} & 0.993/0.467 & 0.991/0.419 & 0.992/0.456 & -- & 0.993/0.495 \\
& Tile & 0.956/0.690 & 0.951/0.597 & 0.902/0.487 & \textbf{0.972}/\textbf{0.688} & -- & 0.931/0.523 \\
& Wood & \textbf{0.955}/\textbf{0.624} & 0.948/0.495 & 0.935/0.439 & 0.940/0.481 & -- & 0.939/0.477 \\
\hline
\multicolumn{2}{@{}l}{Mean}
& \textbf{0.983/0.645} & 0.975/0.583 & 0.966/0.506 & 0.980/0.573 & 0.979/0.618 & 0.978/0.598 \\
\hline
\end{tabular}%
} 
{\textsuperscript{* Only mean was available from source.}}
\end{table*}

\subsection{Ablation Studies}

We conducted ablation studies to validate both the proposed OmniAD architecture and the effectiveness of the data augmentation strategies.

\subsubsection{OmniAD Architecture Ablation}
To validate the architectural changes proposed in OmniAD and quantify the contribution of each key modification compared to the baseline Reverse Distillation  \cite{Deng2022AnomalyEmbedding}, we conducted an ablation study on both OmniAD's components and data augmentation effects. To evaluate the component wise effect of OmniAD we shall refer to the ablation configurations as:

\begin{itemize}
    \item \textbf{OmniAD}: Full model (ResNeXt backbone + $A^i$).
    \item \textbf{OmniAD w/o R}: Original Reverse Distillation  backbone + $A^i$ (isolates $A^i$ contribution without ResNeXt).
    \item \textbf{OmniAD w/o $A^i$}: ResNeXt backbone without $A^i$ (isolates ResNeXt contribution without $A^i$).
    \item \textbf{RD}: Baseline Reverse Distillation model (without $A^i$ and ResNeXt).
\end{itemize}

Results for these configurations are presented in Tables \ref{tab:toycity_1_2}, \ref{tab:toycity_3_4}, \ref{tab:auroc_toycity_1_2}, \ref{tab:auroc_toycity_3_4}, \ref{tab:f1_mad_combined}, and \ref{tab:auroc_mad_combined}. We aggregate the effects of each component of OmniAD without augmentation in Table \ref{tab:f1_avg_summary_simple}.

\begin{table}[htbp] 
  \centering
  \caption{Average Pixel $F_1$ Scores across ToyCity and MAD-Real\cite{ZhouPad:Detection} with \textbf{no augmentation} and percentage improvement over baseline RD.}
  \label{tab:f1_avg_summary_simple}
  \begin{tabular}{lcc}
    \hline 
    \textbf{Method} & \textbf{$F_1$} & \textbf{ Improvement over RD} \\
    \hline 
    OmniAD           & 0.398 & 32.5\% \\
    OmniAD w/o R     & 0.376 & 25.1\% \\
    OmniAD w/o $A^i$ & 0.334 & 11.1\% \\
    RD\cite{Deng2022AnomalyEmbedding} & 0.300 & 0.0\%  \\
    \hline 
  \end{tabular}
\end{table}

\textbf{Contribution of Student Attention Modules ($A^i$):} By comparing configurations with and without $A^i$ while keeping the backbone consistent (OmniAD w/o R vs. RD), we observe the impact of the student attention module. As shown in \tablename~\ref{tab:f1_avg_summary_simple}, the inclusion of $A^i$ leads to a 25.1\% gain over RD. This supports our hypothesis that expanding the ERF via the student attention modules significantly enhances the model's ability to localize anomalies.

\textbf{Contribution of ResNeXt Backbone:} The impact of the ResNeXt backbone is assessed by comparing OmniAD and RD configurations. As shown in \tablename~\ref{tab:f1_avg_summary_simple}, by comparing different backbones (OmniAD vs. OmniAD w/o $A^i$), using ResNeXt provides an 11.1\% gain.

In summary, the architecture ablation study quantitatively demonstrates that both the ResNeXt backbone and the proposed student attention modules ($A^i$) contribute positively to the performance of OmniAD. The combination of these components in OmniAD leads to the superior performance observed in our experiments over Reverse Distillation.

\subsubsection{NVS Augmentation Ablation}
To determine the effectiveness of our Novel View Synthesis (NVS) data augmentation techniques, specifically Query-Aligned Non-anomalous Views (QANV) and Interpolated Non-anomalous Views (INV), we performed an ablation study on OmniAD's performance. The underlying assumption was that augmenting the training data with synthesized non-anomalous views, either aligned with query images or interpolated between existing views, would enhance the model's ability to generalize across different viewpoints and consequently improve anomaly detection. \tablename~\ref{tab:omniad_aug_avg_f1} presents a summary of OmniAD's average pixel $F_1$ scores across all datasets under different augmentation conditions.


\begin{table}
  \centering
  \caption{Average $F_1$ score across ToyCity and MAD-Real\cite{ZhouPad:Detection} for different augmentation strategies.}
  \label{tab:omniad_aug_avg_f1}
  \begin{tabular}{llcc}
    \hline 
    \textbf{Method} & \textbf{Augmentation Strategy} & \textbf{$F_1$} & \textbf{Improvement} \\
    \hline 
    OmniAD & Both   & 0.493 & 64.33\% \\
    OmniAD & QANV   & 0.493 & 64.33\% \\
    OmniAD & INV    & 0.460 & 53.33\% \\
    OmniAD & No Aug & 0.398 & 32.67\% \\
    RD\cite{Deng2022AnomalyEmbedding} & No Aug & 0.300 & 0\% \\
    \hline 
  \end{tabular}
\end{table}

The results in Table~\ref{tab:omniad_aug_avg_f1} clearly indicate that NVS-based augmentation markedly boosts OmniAD's average performance. Both the QANV strategy and the combined use of QANV and INV (Both) achieved the highest average $F_1$ score of 0.493, which is approximately a 64.33\% relative increase compared to Reverse Distillation  without augmentation. The INV strategy, when used alone, also demonstrated a significant improvement, yielding an average $F_1$ score of 0.460.

The detailed results (Tables~\ref{tab:toycity_1_2}, \ref{tab:toycity_3_4}, \ref{tab:f1_mad_combined}) reveal that the best performing augmentation (QANV, INV, or Both) can vary depending on the specific dataset and metric of interest. QANV frequently is shown to be the optimal in terms of $F_1$ for a specific dataset, while INV and "Both" are frequently better in terms of AUROC. However, in real-world applications of Scene AD, selecting the empirically optimal augmentation strategy per dataset is not feasible due to the absence of ground truth labels. Fortunately, combining both QANV and INV offers a robust strategy that does not require dataset-specific optimization. The aggregated results across datasets shown in \tablename~\ref{tab:omniad_aug_avg_f1} shows that the combination of QANV and INV achieves the same high average $F_1$ score as QANV across all datasets. Thus, while combining QANV and INV may not be the most optimal for a specific dataset, the combination is optimal on average.

\section{Limitations and Future Work}
\label{sec:limitations}
OmniAD and the proposed ToyCity dataset address unsupervised Scene Anomaly Detection in multi-view scenarios, yet several limitations remain. First, each ToyCity image contains at most one anomaly at a time, even though multiple anomalies were added and removed from each scene. In contrast, real-world settings such as disaster assessments or industrial inspections may involve multiple anomalies within a single capture. Note that ToyCity does include partial occlusions, and OmniAD itself does not strictly limit detection to a single anomaly region. Second, although ToyCity was captured with real cameras at varying angles and lighting conditions, its tabletop-scale scenes do not fully encompass the complexities of large-scale infrastructure.

Furthermore, our data augmentation assumes sufficient coverage for camera localization. In highly variable environments or when dealing with reflective surfaces, generating high-quality synthetic views can be challenging without accurate camera localization. Nonetheless, as shown in our experiments, OmniAD's performance does not hinge solely on these augmented views.

In future work, we plan to:
\begin{itemize}
    \item Extend ToyCity to include multiple anomalies in a single image and introduce dynamic elements such as moving vehicles or pedestrians.
    \item Validate OmniAD on larger real-world datasets (e.g., complete bridge or ship inspections) to further assess its robustness.
    \item Investigate faster 3D reconstruction, localization, and novel view synthesis methods.
\end{itemize}

\section{Conclusion}
\label{sec:conclusion}
This paper introduced and addressed the task of Scene Anomaly Detection (Scene AD), focusing on localizing anomalies within multi-object scenes captured through unlabeled images exhibiting significant, arbitrary viewpoint variations. We identified the limitations of existing datasets and methods, which typically assume fixed cameras or single-object focus. To bridge this gap, we presented ToyCity, the first multi-object, multi-view real-image dataset specifically designed to evaluate unsupervised AD methods for Scene AD. Furthermore, we demonstrated that established unsupervised AD techniques struggle with pixel-level localization on this challenging dataset. We proposed OmniAD, a novel anomaly detection architecture that refines the Reverse Distillation framework. By incorporating a more robust ResNeXt backbone and, critically, introducing student attention modules to expand the student decoder's effective receptive field (ERF), OmniAD significantly enhances sensitivity to localized anomalies across varying perspectives. Complementing the architecture, we developed two augmentation strategies, Interpolated Non-anomalous Views (INV) and Query-Aligned Non-anomalous Views (QANV), which utilize a Neural Radiance Field for novel view synthesis (NVS) and hierarchical localization to generate targeted non-anomalous training views. Extensive experiments and ablation studies conducted on our ToyCity dataset and the real image subset of MAD validated our approach, showing that OmniAD with QANV and INV improves pixel-level anomaly localization (\( F_1 \) scores) by 64.33\% over Reverse Distillation without QANV and INV. The results underscored the distinct contributions of the ERF expansion via student attention modules and the effectiveness of NVS-based augmentations. We hope our work provides a valuable initial benchmark for future research in view invariant anomaly detection for real-world scenes.

\section{Acknowledgments}
\label{sec:ack}
The authors acknowledge partial financial support from the Texas Department of Transportation (Grant No. 0-7181), and Department of Defense (Project No. G0511607). The contents of this paper reflect the views of the authors, who are responsible for the facts and the accuracy of the data presented herein. The contents do not necessarily reflect the official views or policies of the sponsors. The authors acknowledge the use of the Carya Cluster and the advanced support from the Research Computing Data Core at the University of Houston to carry out the research presented here.

\bibliographystyle{elsarticle-num}
\bibliography{references}






\end{document}